%% file: powmix.tex
\documentclass[10pt,journal,compsoc]{IEEEtran}

\ifCLASSOPTIONcompsoc
  \usepackage[nocompress]{cite}
\else
  \usepackage{cite}
\fi
\ifCLASSINFOpdf
   \usepackage[pdftex]{graphicx}
\else
\fi
\usepackage{amsthm,amsmath,amssymb}
\usepackage{mathrsfs}
\usepackage{booktabs}
\usepackage{multirow}
\usepackage{multicol}
\usepackage{xspace}
\usepackage{bm}
\usepackage[dvipsnames,svgnames,x11names,table]{xcolor}
\usepackage[table]{xcolor}
\usepackage{hyperref}

\usepackage{graphicx}
\usepackage{subcaption}
\usepackage{tikz}
\usetikzlibrary{patterns}
\usepackage{pgfplots}
\pgfplotsset{compat=newest}
\usetikzlibrary{pgfplots.groupplots}

\makeatletter
\renewcommand\paragraph{\@startsection{paragraph}{4}{\z@}{1.5ex}{-1em}{\normalfont\normalsize\bfseries}}
\makeatother

\usepackage{roboto}
\usepackage{pifont}

\newcommand{\circlemarker}[1]{\textcolor{#1}{\Large\textbullet}}

\usetikzlibrary{patterns.meta}
\pgfdeclarepattern{
  name=hatch,
  parameters={\hatchsize,\hatchangle,\hatchlinewidth},
  bottom left={\pgfpoint{-.1pt}{-.1pt}},
  top right={\pgfpoint{\hatchsize+.1pt}{\hatchsize+.1pt}},
  tile size={\pgfpoint{\hatchsize}{\hatchsize}},
  tile transformation={\pgftransformrotate{\hatchangle}},
  code={
    \pgfsetlinewidth{\hatchlinewidth}
    \pgfpathmoveto{\pgfpoint{-.1pt}{-.1pt}}
    \pgfpathlineto{\pgfpoint{\hatchsize+.1pt}{\hatchsize+.1pt}}
    \pgfpathmoveto{\pgfpoint{-.1pt}{\hatchsize+.1pt}}
    \pgfpathlineto{\pgfpoint{\hatchsize+.1pt}{-.1pt}}
    \pgfusepath{stroke}
  }
}

\tikzset{
  hatch size/.store in=\hatchsize,
  hatch angle/.store in=\hatchangle,
  hatch line width/.store in=\hatchlinewidth,
  hatch size=5pt,
  hatch angle=0pt,
  hatch line width=.5pt,
}

\tikzset{every mark/.append style={solid}}
\pgfplotsset{
	grid=both, width=\columnwidth, try min ticks=5,
	every axis/.append style={font=\small},
	every axis plot/.append style={thick,mark=none,mark size=1.5,tension=0.18},
	legend cell align=left, legend style={fill opacity=0.8},
	nodes near coords math/.style={
		nodes near coords={\pgfmathprintnumber[assume math mode=true]{\pgfplotspointmeta}},
	},
	every non boxed x axis/.style={},
}

\usepackage[ruled,vlined,linesnumbered]{algorithm2e}
\input{tex/abbrev}
\input{tex/defn}

\begin{document}

\title{\ours: A Versatile Regularizer for \\ Multimodal Sentiment Analysis}


%
%

\author{Efthymios~Georgiou,~\IEEEmembership{Graduate Student Member,~IEEE,}
        Yannis~Avrithis,~\IEEEmembership{Senior Member,~IEEE,}
        and~Alexandros~Potamianos,~\IEEEmembership{Fellow,~IEEE}%
\IEEEcompsocitemizethanks{
\IEEEcompsocthanksitem Efthymios Georgiou is with the School of ECE, National Technical University of Athens, Athens, Greece, and the Institiute for Speech and Language Processing, Athena Research Center, Athens, Greece \protect\\
E-mail: efthygeo@mail.ntua.gr
\IEEEcompsocthanksitem Yannis Avrithis is with the Institute of Advanced Research on Artificial Intelligence (IARAI), Vienna, Austria \protect\\
E-mail: yannis@avrithis.net
\IEEEcompsocthanksitem Alexandros Potamianos is with the School of ECE, National Technical University of Athens, Athens, Greece \protect\\
E-mail: potam@central.ntua.gr
}
\thanks{Manuscript received xxx; revised xxx. (Corresponding author: Efthymios Georgiou)}
}

%
%


\markboth{Journal of \LaTeX\ Class Files,~Vol.~14, No.~8, August~2015}%
{Shell \MakeLowercase{\textit{et al.}}: Bare Demo of IEEEtran.cls for Computer Society Journals}

\IEEEtitleabstractindextext{%
\begin{abstract}
\emph{Multimodal sentiment analysis} (MSA) leverages heterogeneous data sources to interpret the complex nature of human sentiments. Despite significant progress in multimodal architecture design, the field lacks comprehensive regularization methods. This paper introduces \ours, a versatile embedding space regularizer that builds upon the strengths of unimodal mixing-based regularization approaches and introduces novel algorithmic components that are specifically tailored to multimodal tasks. \ours is integrated before the fusion stage of multimodal architectures and facilitates intra-modal mixing, such as mixing text with text, to act as a regularizer.
\ours consists of five components:
1) a varying number of generated mixed examples,
2) mixing factor reweighting,
3) anisotropic mixing,
4) dynamic mixing, and
5) cross-modal label mixing.
Extensive experimentation across benchmark MSA datasets and a broad spectrum of diverse architectural designs demonstrate the efficacy of \ours, as evidenced by consistent performance improvements over baselines and existing mixing methods. An in-depth ablation study highlights the critical contribution of each \ours component and how they synergistically enhance performance. Furthermore, algorithmic analysis demonstrates how \ours behaves in different scenarios, particularly comparing early versus late fusion architectures. Notably, \ours enhances overall performance without sacrificing model robustness or magnifying text dominance. It also retains its strong performance in situations of limited data. Our findings position \ours as a promising versatile regularization strategy for MSA. Code will be made available.
\end{abstract}


}


\maketitle

\IEEEdisplaynontitleabstractindextext

%
\IEEEpeerreviewmaketitle


\input{tex/intro}
\input{tex/related}
\input{tex/back}
\input{tex/method}

\input{tex/exp-setup}
\input{tex/exp-results}

\input{tex/exp-ablation}

\input{tex/exp-analysis}
\input{tex/conclusion}



%







\ifCLASSOPTIONcaptionsoff
  \newpage
\fi



\bibliographystyle{IEEEtran}
\bibliography{powmix}
\end{document}

%% file: tex/abbrev.tex
\newcommand{\alert}[1]{{\color{red}{#1}}}

\newcommand{\eq}[1]{(\ref{eq:#1})}

\newcommand{\Th}[1]{\textsc{#1}}
\newcommand{\mr}[2]{\multirow{#1}{*}{#2}}
\newcommand{\mc}[2]{\multicolumn{#1}{c}{#2}}
\newcommand{\tb}[1]{\textbf{#1}}
\newcommand{\ch}{\checkmark}

\newcommand{\citeme}[1]{\alert{[X]}}
\newcommand{\refme}[1]{\alert{(X)}}

\newcommand{\tran}{^\top}

\newcommand{\real}{\mathbb{R}}

\newcommand{\defn}{\mathrel{:=}}

\newcommand{\wt}[1]{\widetilde{#1}}

\newcommand{\cD}{\mathcal{D}}

\newcommand{\cM}{\mathcal{M}}

\newcommand{\cP}{\mathcal{P}}

\newcommand{\cX}{\mathcal{X}}

\newcommand{\vH}{\mathbf{H}}

\newcommand{\vM}{\mathbf{M}}

\newcommand{\vP}{\mathbf{P}}

\newcommand{\vX}{\mathbf{X}}

\newcommand{\va}{\mathbf{a}}

\newcommand{\vh}{\mathbf{h}}

\newcommand{\vy}{\mathbf{y}}

\newcommand{\vone}{\mathbf{1}}

\newcommand{\valpha}{{\boldsymbol{\alpha}}}

\newcommand{\rLambda}{\mathrm{\Lambda}}

\newcommand{\vLambda}{\bm{\rLambda}}

\makeatletter
\newcommand*\bdot{\mathpalette\bdot@{.7}}
\newcommand*\bdot@[2]{\mathbin{\vcenter{\hbox{\scalebox{#2}{$\m@th#1\bullet$}}}}}
\makeatother

\makeatletter
\DeclareRobustCommand\onedot{\futurelet\@let@token\@onedot}
\def\@onedot{\ifx\@let@token.\else.\null\fi\xspace}

\def\eg{\emph{e.g}\onedot} 
\def\ie{\emph{i.e}\onedot} 
 
\def\etc{\emph{etc}\onedot} 
\def\wrt{w.r.t\onedot}  
 
\def\etal{\emph{et al}\onedot}
\makeatother

%% file: tex/defn.tex
\newcommand{\Bern}{\operatorname{Bern}}
\newcommand{\Beta}{\operatorname{Beta}}
\newcommand{\Dir}{\operatorname{Dir}}

\newcommand{\ours}{$\cP$owMix\xspace}

\definecolor{Gray0}{gray}{0.4}
\definecolor{Gray1}{gray}{0.75}
\definecolor{Gray2}{gray}{0.80}
\definecolor{Gray2}{gray}{0.82}
\definecolor{Gray3}{RGB}{205, 246, 250} 
\definecolor{Gray4}{gray}{0.95}
\definecolor{my_Green}{RGB}{0,140,0}

\definecolor{LightBlue1}{RGB}{173, 216, 230} 
\definecolor{LightBlue2}{RGB}{135, 206, 250} 
\definecolor{LightBlue3}{RGB}{176, 224, 230} 

\newcommand{\rd}[1]{\textcolor{red}{#1}}

\definecolor{CB91_Blue}{HTML}{2CBDFE}
\definecolor{CB91_Green}{HTML}{47DBCD}
\definecolor{CB91_Pink}{HTML}{F3A0F2}
\definecolor{CB91_Purple}{HTML}{9D2EC5}
\definecolor{CB91_Violet}{HTML}{661D98}
\definecolor{CB91_Amber}{HTML}{F5B14C}

\definecolor{Point_Color}{HTML}{9b59b6}
\definecolor{Hull_Line}{HTML}{34495e}
\definecolor{Combination}{HTML}{e67e22}

%% file: tex/intro.tex
\IEEEraisesectionheading{
\section{Introduction}
\label{sec:introduction}
}

%
%
%
%

\IEEEPARstart{M}{}ultimodal Sentiment Analysis (MSA) is the task of enriching a computer system with affective understanding of real-world human-centric video segments. Interpreting sentiments from videos is very challenging due to the multifaceted nature of human communication through speech, facial expressions, linguistic content, etc.~\cite{narayanan_2013_behavioral_signal}. The practical applications of MSA are numerous in the digital era and vary from \emph{human-computer interaction} (HCI)~\cite{filby_2019_child_robot} and healthcare~\cite{poria2017review}, to opinion mining in reviews~\cite{stappen2021multimodal} and education~\cite{affect_in_education}. Despite the advancements in the MSA field~\cite{metallinou2012context, soleymani2017survey, zadeh2017tensor, tsai2019multimodal, hazarika2020misa, yu2021learning, sun2022learning}, developing an end-to-end system that effectively analyzes the complex aspects of human sentiment remains an open research challenge.

Building on the idea that multimodal learning includes unimodal elements, such as decomposing multimodal predictions into separate unimodal contributions and multimodal interactions~\cite{hessel-lee-2020-multimodal}, we perceive multimodal tasks as being fundamentally more complex than unimodal. This suggests that challenges inherent in unimodal setups, such as overfitting, also exist in multimodal scenarios. In addition, the diverse nature of input data as manifested by the coexistence of the symbolic language modality with lower-level visual cues, further hinders the learning process.

Despite expectations that multimodal networks would outperform their unimodal counterparts~\cite{huang_2021_mm_better}, this is not consistently observed. It is found that different inputs generalize at different rates, leading to unexpected performance degradation~\cite{wang2020makes}, as well as a tendency of networks to over-rely on dominant modalities~\cite{wu2022characterizing}, \eg, text in MSA~\cite{GKOUMAS2021WhatMakesTheDiff}. Furthermore, studies demonstrate that joint multimodal training tends to learn a limited spectrum of features and modalities~\cite{huang_22_mm_fail}, resulting in suboptimal solutions.

Considering the challenges in multimodal learning, it is reasonable to speculate that regularization and data augmentation methods may be beneficial, similar to unimodal tasks. Nevertheless, the existing literature on this topic remains relatively sparse and of limited scope. Some approaches like Wang et al.~\cite{wang2020makes} and Wu et al.~\cite{wu2022characterizing}, propose to dynamically reweight unimodal loss terms within the overall learning objective, while Du et al.~\cite{du2023onunimodal} suggest leveraging a unimodal teacher model to improve learning of unimodal features. However, these methods are tied with specific learning hypotheses and late fusion architectures, which constrains their broader applicability.

For more advanced models, like those employed for MSA, Liu et. al.~\cite{liu2023learning} introduce a learnable auto-encoder for embedding augmentation within multimodal networks, and M$3$~\cite{georgiou21_interspeech} utilizes intense text masking in the latent unimodal space before fusion, acting as a regularizer. More closely aligned with our work, AV-MC~\cite{liu2022_sims2} employs MixUp~\cite{zhang2018mixup} independently for acoustic and visual streams, when labels are available for each modality, requiring three separate forward propagations, one for the original input and one for each unimodal set of mixed examples. However, all these approaches are bound to specific architectural designs, fusion strategies and learning assumptions, which restrict their applicability. Therefore, a broad-spectrum regularization framework that transcends such constraints is crucial for multimodal learning environments such as MSA.

In this work, we introduce \ours, a novel regularization method specifically crafted to improve regularization in multimodal scenarios and in particular MSA. Unlike methods designed to handle modality-specific challenges, \ours aims to offer a broad-spectrum solution applicable across a range of datasets and model architectures. As a member of the mixing algorithm family, it is inspired by methods like MixUp~\cite{zhang2018mixup},  TransMix~\cite{transmix} and MultiMix~\cite{venkataramanan2023Embedding}, known for their regularization capabilities. \ours is integrated before the fusion stage in the multimodal architecture and facilitates intra-modal mixing, \eg, text with text and audio with audio.

What sets \ours apart is its novel components that, in synergy, render it suitable for multimodal contexts. In particular, the algorithm is characterized by five key features: 1) a \emph{varying number of generated mixed examples}, 2) \emph{mixing factor reweighting} to encapsulate the importance of each modality for each mini-batch example, 3) \emph{anisotropic mixing} for independent mixing across unimodal spaces, 4) \emph{dynamic mixing}, a novel element for representation mixing, and 5) \emph{cross-modal label mixing}, a new way to aggregate mixed labels across modalities. \ours emerges as a versatile regularizer across MSA datasets and architectures. To the best of our knowledge, such an approach has not been previously reported in the literature.

To establish the effectiveness of \ours, we conduct experiments on three widely used MSA benchmark datasets: MOSI~\cite{zadeh2016multimodal}, MOSEI~\cite{zadeh2018multimodal}, and SIMS~\cite{yu2020ch}. We also employ three different multimodal architectures: MulT~\cite{tsai2019multimodal}, MISA~\cite{hazarika2020misa}, and Self-MM~\cite{yu2021learning}. These models are chosen because they perform well and, most importantly, cover a wide range of architectural designs and learning approaches.

Our main contributions are summarized as follows:
\begin{enumerate}
	\item We introduce \ours, a novel regularization method applied to MSA. It consists of five key components, two of which build upon existing ideas and the other three are entirely novel: anisotropic mixing, dynamic mixing, and cross-modal label mixing.
	\item We experimentally validate the effectiveness of \ours across diverse MSA datasets and architectures, confirming superior performance over baselines and state-of-the-art mixing methods.
	\item We conduct an in-depth ablation study of the features of \ours, demonstrating the contribution of each component to the overall performance. We also highlight the synergetic impact of these features.
	\item We present a comprehensive algorithmic analysis, demonstrating the behavior of \ours across different fusion types, its robustness to noise and text dominance levels, as well as its efficacy under limited data scenarios.
\end{enumerate}

The paper is structured as follows: \autoref{sec:related_work} covers related work and \autoref{sec:background} provides formulation and background for our study. Next, \autoref{sec:method} details the \ours algorithm, while \autoref{sec:exp_setup} outlines our experimental setup. The core experimental results, as well as the ablation study and algorithmic analysis, are presented in \autoref{sec:experiments}. Finally, \autoref{sec:conclusion} draws conclusions and discusses future research directions.

%% file: tex/related.tex
\section{Related Work}
\label{sec:related_work}

This section provides an overview of the literature, beginning with an exploration of works in the MSA field, which is the core of our experimentation. We then discuss advancements in mixing techniques within the unimodal learning context, highlighting their components. Finally, we present existing multimodal regularizers and highlight their task and problem specific characteristics.


\subsection{Multimodal sentiment analysis}

MSA research mainly focuses on building better fusion schemes and utilizing diverse learning recipes to enhance representation learning for the task at hand. In particular, TFN~\cite{zadeh2017tensor} employs outer product of unimodal representations to capture cross-modal interactions. Poria \etal~\cite{poria2017multi} and Gu \etal~\cite{gu2018multimodal} implement multi-level and hierarchical attention to better contextualize information. DHF~\cite{georgiou2019deep} applies a hierarchical fusion mechanism across different levels within the architecture.

Other types of neural structures employed in MSA include neural memory modules~\cite{zadeh2018multimodal}, capsule networks~\cite{tsai20routing}, and graph neural networks~\cite{joshi-etal-2022-cogmen}. Tsai \etal~\cite{tsai2019multimodal} utilize transformers, where cross-attention blocks act as early fusion and concatenation serves as late fusion. Rahman \etal~\cite{rahman2020integrating} fine-tune a pre-trained BERT~\cite{devlin2019bert} model by incorporating a multimodal shifting layer as early fusion.

Another line of work utilizes more complex learning recipes such as canonical correlation analysis~\cite{sun2022learning} and cycle-consistency loss~\cite{pham2019found} across modalities. Coupling different learning recipes with pre-trained models has been a popular choice among researchers. Yu \etal~\cite{yu2021learning} introduce a unimodal pseudo-labeling module that backpropagates three additional losses. Hazarika \etal~\cite{hazarika2020misa} augment the learning objective with feature reconstruction loss as well as attracting and repelling objectives.

A two-step hierarchical learning recipe based on mutual information maximization is proposed in~\cite{han2021improving}, while Sun \etal~\cite{sun2022learning} propose a meta-learning framework that learns each unimodal network and then adapts them for the MSA task. Sun \etal~\cite{sun2023efficient} propose a transformer architecture leveraging dual-level reconstruction loss and an attraction loss in a Siamese setup between complete and incomplete data. Hu \etal~\cite{hu-etal-2022-unimse} employ a text encoder-decoder architecture, using T5~\cite{RAFFEL2020T5}, and implement a contrastive loss among unimodal encoders. The decoder generates text sequences, which are decoded into MSA-related info such as polarity. Notably, none of the aforementioned approaches handles multimodal regularization.


\subsection{Mixing in unimodal learning setups}

Regularization\footnote{In this work, we attribute regularization as any modification we make to a learning algorithm that is intended to reduce its generalization error but not its training error~\cite{Goodfellow-et-al-2016}.} in unimodal learning setups has been extensively studied. We primarily focus on techniques that modify the learning process through mixing-based algorithms such as MixUp~\cite{zhang2018mixup}. These algorithms are notable for their regularization benefits and a unique capability to mix representations in the latent space. This feature is particularly desired for the development of broad multimodal regularizers.

\paragraph*{Input mixing}

Algorithmic approaches in this category are usually attached to specific types of data. For \emph{computer vision} (CV), the most studied field, options have evolved from fundamental transformations, \eg, translation and rotation~\cite{Goodfellow-et-al-2016}, to more advanced methods. These include MixUp~\cite{zhang2018mixup} and CutMix~\cite{yun2019cutmix}, which mix pairs of images in the pixel and label space, as well as AutoMix~\cite{liu2022automix} that introduces an automatic mixing framework. For a unified study on vision mixing techniques, we refer to~\cite{li2022openmixup}.

In \emph{natural language processing} (NLP), mixing words in their raw format is not straightforward, leading to approaches like SSMix~\cite{yoon2021ssmix}, which substitutes salient parts of a sentence with words from another. In the Audio and Speech domain, SpecMix~\cite{kim21c_interspeech} mixes two spectrogram representations \wrt the frequency domain. Our work shares a similar idea with TransMix~\cite{transmix}, a state-of-the-art CV technique, proposing pixel-wise reweighting of mixing factors based on their attention map values.

\paragraph*{Latent space mixing}

Algorithms in this category focus on manipulating latent representations. Manifold MixUp~\cite{verma2019manifold} interpolates pairs of hidden representations along with their labels. Non-Linear MixUp~\cite{Guo_2020_nonlinear} extends this concept with a non-linear interpolation scheme in the text embedding space. ReMix~\cite{chou_2020_remix} favors the minority class during mixing. Further expanding on these ideas, SpeechMix~\cite{jindal20_speechmix} and MixUp-Transformer~\cite{sun_2020_mixuptrans}, are variants of Manifold MixUp for speech and NLP tasks respectively.

Closely related to our work is MultiMix~\cite{venkataramanan2023Embedding}, a state-of-the-art method that mixes all representations within a batch to generate many mixtures. MultiMix also incorporates the idea of reweighting interpolation factors prior to mixing. Similarly, \ours proposes a methodology to generate more mixed examples than the mini-batch size, yet by interpolating fewer examples than MultiMix. These algorithms, especially MultiMix, represent a methodological shift towards more abstract regularization methods and motivate the need for techniques beyond unimodal boundaries.


\subsection{Regularization in multimodal setups}
\label{sec:joint_multimodal_reg}

Here, we position the proposed \ours in the context of existing multimodal learning regularization techniques. Current methods often target specific challenges~\cite{wang2020makes, du2023onunimodal} or are confined to particular inputs and domains. For example, MixGen~\cite{Hao_2023_WACV} combines image and text data intra-modally but is primarily helpful for tasks like retrieval and captioning. Similarly, \emph{cross modal CutMix} (CMC)~\cite{wang_22_cmc} bridges unpaired image-text datasets cross-modally but has limited architecture  and task flexibility. By contrast, \ours is designed to adapt across a wide range of supervised multimodal classification or regression problems without being limited to specific input types or architectures.

The embedding space augmentation technique proposed in~\cite{liu2023learning} marks progress in multimodal regularization. However, its complexity limits its application, involving additional learnable parameters, doubled forward propagations, adversarial-like optimization, and manual output thresholding. Similarly, AV-MC~\cite{liu2022_sims2}, which utilizes MixUp for acoustic and visual streams independently, requires unimodal labels and three forward propagations. \ours, on the other hand, offers versatility and broad applicability by not posing any constraints on the learning setup, particularly in complex tasks like MSA.

%% file: tex/back.tex
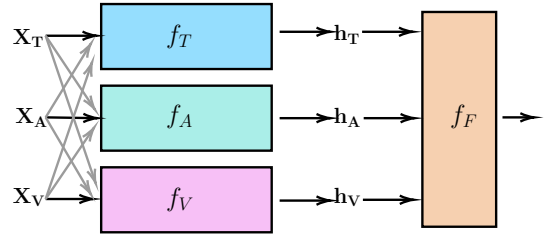
\begin{figure}
\centering
\scalebox{.4}{\input{fig/late_fusion.tikz}}
\caption{\emph{Abstract multimodal fusion scheme} for MSA architectures including MulT~\cite{tsai2019multimodal}, MISA~\cite{hazarika2020misa} and Self-MM~\cite{yu2021learning}. Independent processing pathways for each modality, represented by \emph{encoders} $f_m$, where $m \in \{T, A, V\}$ is the modality ($T$: text, $A$: acoustic, $V$: video). In models like MulT, encoders can incorporate other modalities as inputs too, \ie, early-fusion. The hidden representations $\vh_m$ extracted by the encoders are fed to the \emph{fusion network} $f_F$, which generates the final prediction. Depending on the architecture, $f_F$ can manifest as a non-linear feedforward network (MulT), a single-layer transformer (MISA), a dual linear layer setup (Self-MM), \etc. This scheme abstracts away components of the architecture not directly related to the prediction task. Mixing is performed directly on $\vh_m$.}
\label{fig:abstract_fusion}
\end{figure}

\section{Background}
\label{sec:background}

We formulate \emph{multimodal sentiment analysis} (MSA) as a multimodal fusion task and present an abstract architecture scheme that encapsulates most existing approaches, as illustrated in \autoref{fig:abstract_fusion}. We then describe unimodal mixing algorithms with a particular focus on Manifold MixUp~\cite{verma2019manifold} and MultiMix~\cite{venkataramanan2023Embedding}. We also briefly outline the idea of \emph{reweighting}, a mechanism used in different mixing algorithms~\cite{transmix}.

\begin{figure*}
\centering
\begin{subfigure}{0.24\textwidth}
	\centering
	\includegraphics[width=\textwidth]{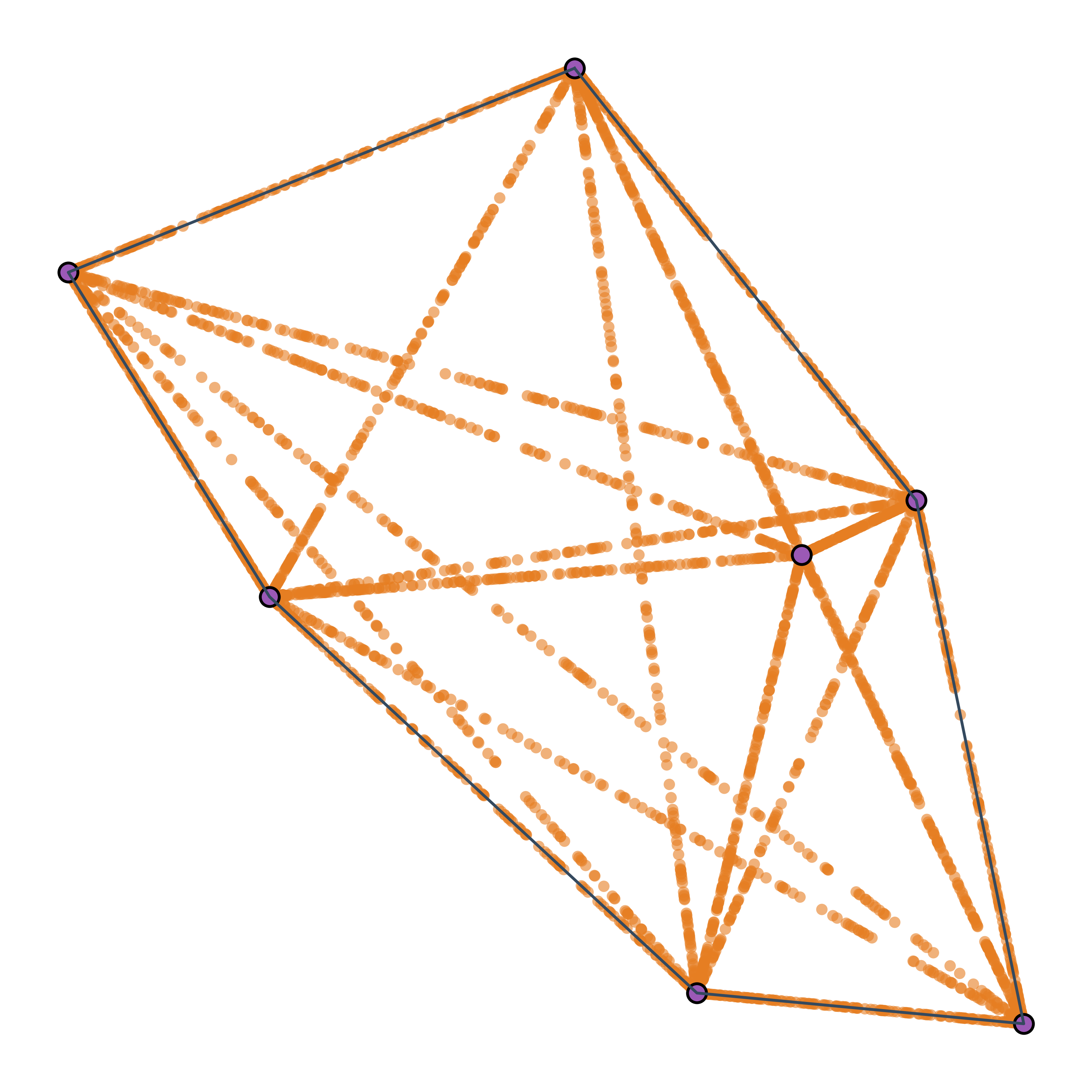}
	\caption{$n_I=2$ (MixUp)}
        \label{fig:mixup}
\end{subfigure}
\hfill
\begin{subfigure}{0.24\textwidth}
	\centering
	\includegraphics[width=\textwidth]{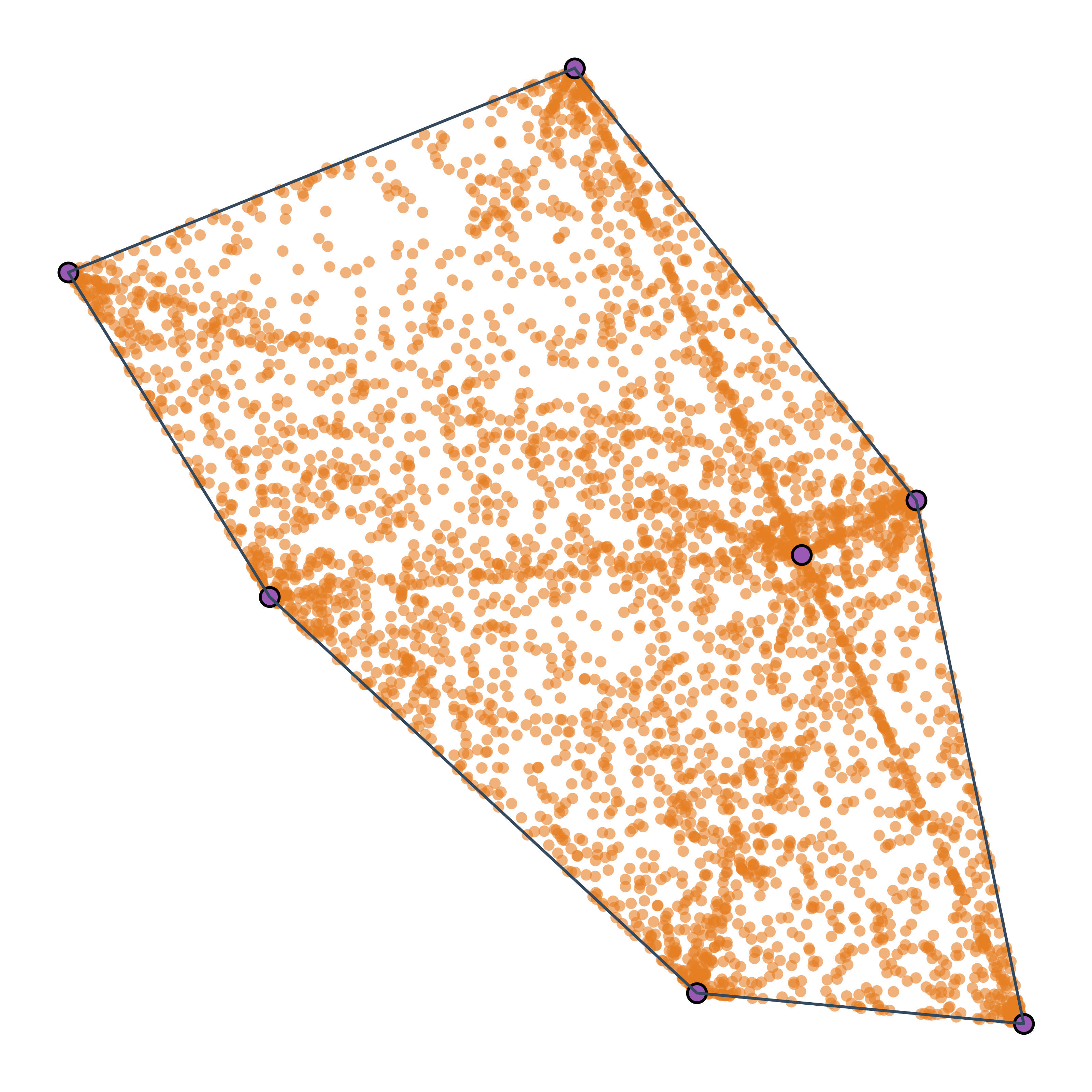}
	\caption{$n_I=3$}
	\label{fig:powmix_3}
\end{subfigure}
\hfill
\begin{subfigure}{0.24\textwidth}
	\centering
	\includegraphics[width=\textwidth]{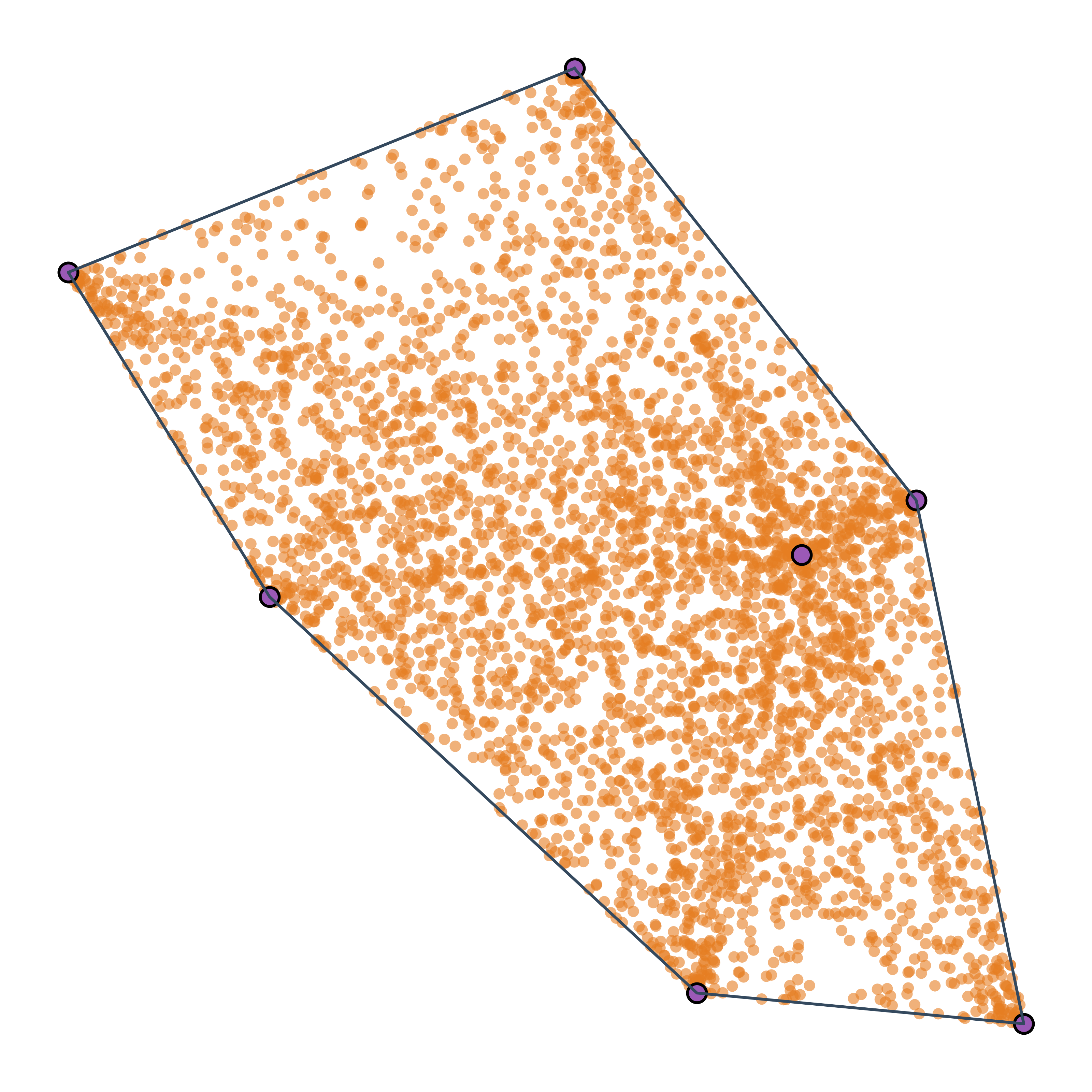}
	\caption{$n_I=4$}
	\label{fig:powmix_4}
\end{subfigure}
\hfill
\begin{subfigure}{0.24\textwidth}
	\centering
	\includegraphics[width=\textwidth]{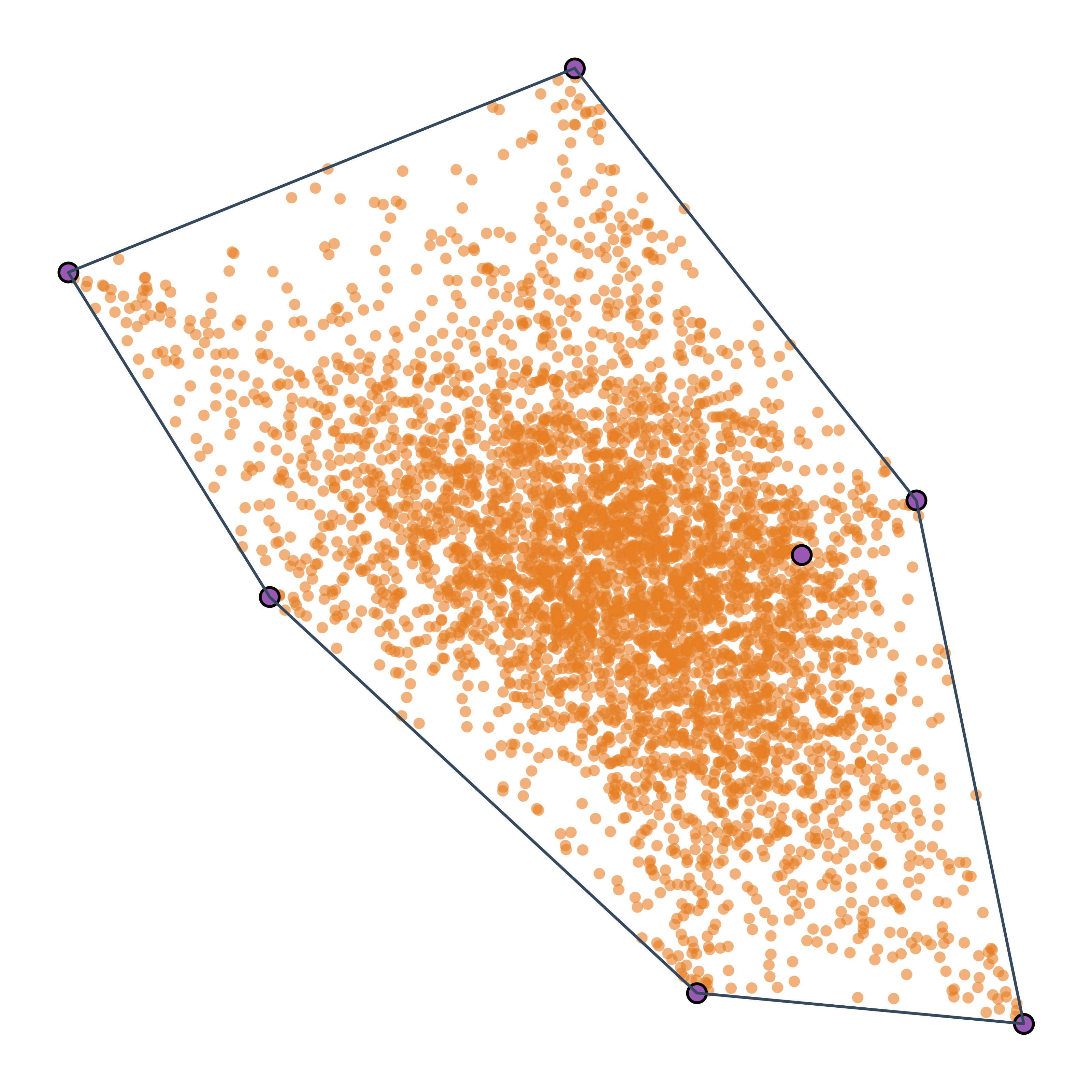}
	\caption{$n_I = 7$ (MultiMix)}
	\label{fig:multimix}
\end{subfigure}
\caption{Motivation of \ours. Given seven 2-dimensional points, we generate $2^{12}$ points, each as a convex combination of a random subset of $n_I$ points. From left to right, we interpolate pairs, triplets, quadruplets, and all seven points. MixUp (a) interpolates two points at a time, while MultiMix (d) uses all seven points. By \emph{dynamic mixing}, \ours randomly samples a subset of different cardinality for each generated point, hence can provide mixing instances between interpolation of pairs and all points. Empirically, we find that mixing from $2$ to $4$ points gives good performance.
\circlemarker{Point_Color}: Original examples;
\circlemarker{Combination}: interpolated examples;
\textcolor{Hull_Line}{\rule[0.5ex]{2em}{1.5pt}}: convex hull.}
\label{fig:n_I_illustation}
\end{figure*}

\subsection{Preliminaries and notation}

\autoref{fig:abstract_fusion} shows an abstract multimodal fusion architecture used in MSA models, \eg, ~\cite{zadeh2017tensor, tsai2019multimodal, hazarika2020misa, yu2021learning, georgiou21_interspeech}.
We have omitted parts of the architecture not directly involved in the predictive information flow, such as the decoder in MISA~\cite{hazarika2020misa} and the unimodal label generation module in Self-MM~\cite{yu2021learning}. Our focus is on the input modality encoders $f_m$ and the fusion network $f_F$ alone.

Each unimodal input space is denoted as $\cX_m$, where $m$ is the modality, from a set of indices $\cM = \{1,\dots, M\}$. The multimodal input space is denoted by the Cartesian product $\cX_\cM = \cX_1 \times \dots \times \cX_M$. The multimodal input data is given as the collection $\cD = \{ \vX^{(i)}, y^{(i)} \}_{i=1}^N$, where $\vX^{(i)} = (\vX^{(i)}_1, \dots, \vX^{(i)}_M)$ is a multimodal $m$-tuple with each $\vX^{(i)}_m \in \cX_m$, $y^{(i)} \in \real$ is the corresponding label for regression tasks and $N$ is the number of multimodal $m$-tuples. When discussing unimodal concepts, such as Manifold MixUp, we omit the $m$ subscript.

Each input modality encoder is a mapping $f_m: \cX_\cM \to \real^{d_m}$, where $d_m$ is the hidden space dimension. Given multimodal input $\vX \in \cX_\cM$, the output of each encoder is $\vh_m = f_m(\vX) \in \real^{d_m}$, allowing for early fusion in general. The fusion network is a mapping $f_F:$ $\real^{d_1} \times \dots \times \real^{d_M} \to \real$. The complete prediction scheme $f$ is the composition of input modality mappings with the fusion map, \ie, $f(\vX) = f_F(\vh_1, \dots, \vh_M) = f_F(f_1(\vX), \dots, f_M(\vX)) $. All mixing operations are applied to the hidden representations of an entire mini-batch, denoted by $\vH_m \in \real^{B \times d_m}$, where $B$ is the mini-batch size.


\subsection{Manifold MixUp}

Manifold MixUp~\cite{verma2019manifold} is a unimodal regularization method that interpolates pairs of hidden representations from different input examples using an interpolation factor $\lambda \in [0, 1]$.
Given a pair of representations $\mathbf{h}^{(i)}, \mathbf{h}^{(j)}$ in latent unimodal space $ \real^d$ and their labels $y^{(i)}, y^{(j)}$, mixing is performed by the convex combination
\begin{align}
	\wt{\vh} &= \lambda \vh^{(i)} + (1 - \lambda) \vh^{(j)} \\
	\wt{y} &= \lambda y^{(i)} + (1 - \lambda) y^{(j)}.
\end{align}
Following standard practice, $\lambda$ is sampled from distribution $\Beta(\alpha)$, where $\alpha = 1$. As shown in \autoref{fig:mixup}, this method results in mixed embeddings along linear segments between pairs of examples.


\subsection{MultiMix}
\label{sec:multimix}

MutiMix~\cite{venkataramanan2023Embedding}, a state-of-the-art mixing method, is incorporated in our study for comparative analysis with our proposed \ours. Unlike Manifold MixUp, which interpolates pairs of examples, MultiMix interpolates all examples in a batch and generates a number of mixed examples that is independent of the mini-batch size. In particular, given a batch of size $B$, MultiMix mixes all $B$ hidden representations, denoted by $\vH \in \real^{B\times d}$, to generate $n_O$ mixed examples. This involves randomly sampling $n_O$ interpolation vectors in $\real^B$ from Dirichlet distribution $\Dir(\valpha)$, resulting in an interpolation matrix $\vLambda \in \real^{n_O \times B}$. Mixing is then performed by the convex combination
\begin{align}
	\wt{\vH} &= \vLambda \vH \in \real^{n_O \times d} \label{eq:multimix_h} \\
	\wt{\vy} &= \vLambda \vy \in \real^{n_O}, \label{eq:multimix_y}
\end{align}
where $\vy \in \real^B$ denotes the labels of the mini-batch. Parameter $\alpha$ is sampled from a uniform distribution $U[0.5, 2.0]$ by default, which produces mixed examples as shown in \autoref{fig:multimix}. The number $n_O$ of mixed examples is a tunable hyperparameter. In \ours, we adopt this idea of $n_O$ being decoupled from the mini-batch size.


\subsection{Mixing factor reweighting}
\label{sec:repr_reweight}

Reweighting the mixing factors based on representations has proven an effective strategy in various MixUp variants, such as TransMix and MultiMix~\cite{transmix, venkataramanan2023Embedding}. This idea is founded on the principle that reweighting different elements by assigning them attention-like values can result in more effective mixtures.
The key to this approach is a mapping $g$, which determines the attention weights of mixed elements. This function is based on attention maps obtained by transformer architectures in TransMix~\cite{transmix} and a cross-attention operation over dense features in MultiMix~\cite{venkataramanan2023Embedding}. To keep our approach generic in terms of architecture, we rather use a simpler mapping $g$ in \ours.


%% file: fig/late_fusion.tikz.tex
\tikzset{every picture/.style={line width=0.75pt}} 

\begin{tikzpicture}[x=0.75pt,y=0.75pt,yscale=-1,xscale=1]

\draw  [fill={rgb, 255:red, 44; green, 189; blue, 254 }  ,fill opacity=0.5 ][line width=2.25]  (108,5) -- (323,5) -- (323,87) -- (108,87) -- cycle ;
\draw [line width=2.25]    (38,45) -- (98,45) ;
\draw [shift={(102,45)}, rotate = 180] [color={rgb, 255:red, 0; green, 0; blue, 0 }  ][line width=2.25]    (17.49,-5.26) .. controls (11.12,-2.23) and (5.29,-0.48) .. (0,0) .. controls (5.29,0.48) and (11.12,2.23) .. (17.49,5.26)   ;
\draw  [fill={rgb, 255:red, 71; green, 219; blue, 205 }  ,fill opacity=0.46 ][line width=2.25]  (108,108) -- (323,108) -- (323,190) -- (108,190) -- cycle ;
\draw [line width=2.25]    (38,148) -- (99,148.94) ;
\draw [shift={(103,149)}, rotate = 180.88] [color={rgb, 255:red, 0; green, 0; blue, 0 }  ][line width=2.25]    (17.49,-5.26) .. controls (11.12,-2.23) and (5.29,-0.48) .. (0,0) .. controls (5.29,0.48) and (11.12,2.23) .. (17.49,5.26)   ;
\draw  [fill={rgb, 255:red, 243; green, 160; blue, 242 }  ,fill opacity=0.66 ][line width=2.25]  (108,211) -- (323,211) -- (323,293) -- (108,293) -- cycle ;
\draw [line width=2.25]    (38,251) -- (94,251) ;
\draw [shift={(98,251)}, rotate = 180] [color={rgb, 255:red, 0; green, 0; blue, 0 }  ][line width=2.25]    (17.49,-5.26) .. controls (11.12,-2.23) and (5.29,-0.48) .. (0,0) .. controls (5.29,0.48) and (11.12,2.23) .. (17.49,5.26)   ;
\draw [color={rgb, 255:red, 155; green, 155; blue, 155 }  ,draw opacity=1 ][fill={rgb, 255:red, 155; green, 155; blue, 155 }  ,fill opacity=1 ][line width=2.25]    (38,148) -- (92.91,58.41) ;
\draw [shift={(95,55)}, rotate = 121.5] [color={rgb, 255:red, 155; green, 155; blue, 155 }  ,draw opacity=1 ][line width=2.25]    (17.49,-5.26) .. controls (11.12,-2.23) and (5.29,-0.48) .. (0,0) .. controls (5.29,0.48) and (11.12,2.23) .. (17.49,5.26)   ;
\draw [color={rgb, 255:red, 155; green, 155; blue, 155 }  ,draw opacity=1 ][line width=2.25]    (38,251) -- (96.76,164.31) ;
\draw [shift={(99,161)}, rotate = 124.13] [color={rgb, 255:red, 155; green, 155; blue, 155 }  ,draw opacity=1 ][line width=2.25]    (17.49,-5.26) .. controls (11.12,-2.23) and (5.29,-0.48) .. (0,0) .. controls (5.29,0.48) and (11.12,2.23) .. (17.49,5.26)   ;
\draw [color={rgb, 255:red, 155; green, 155; blue, 155 }  ,draw opacity=1 ][line width=2.25]    (38,251) -- (95.22,82.29) ;
\draw [shift={(96.5,78.5)}, rotate = 108.73] [color={rgb, 255:red, 155; green, 155; blue, 155 }  ,draw opacity=1 ][line width=2.25]    (17.49,-5.26) .. controls (11.12,-2.23) and (5.29,-0.48) .. (0,0) .. controls (5.29,0.48) and (11.12,2.23) .. (17.49,5.26)   ;
\draw [color={rgb, 255:red, 155; green, 155; blue, 155 }  ,draw opacity=1 ][line width=2.25]    (38,45) -- (97.78,134.67) ;
\draw [shift={(100,138)}, rotate = 236.31] [color={rgb, 255:red, 155; green, 155; blue, 155 }  ,draw opacity=1 ][line width=2.25]    (17.49,-5.26) .. controls (11.12,-2.23) and (5.29,-0.48) .. (0,0) .. controls (5.29,0.48) and (11.12,2.23) .. (17.49,5.26)   ;
\draw [color={rgb, 255:red, 155; green, 155; blue, 155 }  ,draw opacity=1 ][line width=2.25]    (38,148) -- (92.97,241.55) ;
\draw [shift={(95,245)}, rotate = 239.56] [color={rgb, 255:red, 155; green, 155; blue, 155 }  ,draw opacity=1 ][line width=2.25]    (17.49,-5.26) .. controls (11.12,-2.23) and (5.29,-0.48) .. (0,0) .. controls (5.29,0.48) and (11.12,2.23) .. (17.49,5.26)   ;
\draw [color={rgb, 255:red, 155; green, 155; blue, 155 }  ,draw opacity=1 ][line width=2.25]    (38,45) -- (99.72,227.21) ;
\draw [shift={(101,231)}, rotate = 251.29] [color={rgb, 255:red, 155; green, 155; blue, 155 }  ,draw opacity=1 ][line width=2.25]    (17.49,-5.26) .. controls (11.12,-2.23) and (5.29,-0.48) .. (0,0) .. controls (5.29,0.48) and (11.12,2.23) .. (17.49,5.26)   ;
\draw [line width=2.25]    (333,40) -- (393,40) ;
\draw [shift={(397,40)}, rotate = 180] [color={rgb, 255:red, 0; green, 0; blue, 0 }  ][line width=2.25]    (17.49,-5.26) .. controls (11.12,-2.23) and (5.29,-0.48) .. (0,0) .. controls (5.29,0.48) and (11.12,2.23) .. (17.49,5.26)   ;
\draw [line width=2.25]    (333,149) -- (393,149) ;
\draw [shift={(397,149)}, rotate = 180] [color={rgb, 255:red, 0; green, 0; blue, 0 }  ][line width=2.25]    (17.49,-5.26) .. controls (11.12,-2.23) and (5.29,-0.48) .. (0,0) .. controls (5.29,0.48) and (11.12,2.23) .. (17.49,5.26)   ;
\draw [line width=2.25]    (333,253) -- (393,253) ;
\draw [shift={(397,253)}, rotate = 180] [color={rgb, 255:red, 0; green, 0; blue, 0 }  ][line width=2.25]    (17.49,-5.26) .. controls (11.12,-2.23) and (5.29,-0.48) .. (0,0) .. controls (5.29,0.48) and (11.12,2.23) .. (17.49,5.26)   ;
\draw  [fill={rgb, 255:red, 230; green, 126; blue, 34 }  ,fill opacity=0.36 ][line width=2.25]  (513,15) -- (606,15) -- (606,286) -- (513,286) -- cycle ;
\draw [line width=2.25]    (614,148) -- (650,148) ;
\draw [shift={(654,148)}, rotate = 180] [color={rgb, 255:red, 0; green, 0; blue, 0 }  ][line width=2.25]    (17.49,-5.26) .. controls (11.12,-2.23) and (5.29,-0.48) .. (0,0) .. controls (5.29,0.48) and (11.12,2.23) .. (17.49,5.26)   ;
\draw [line width=2.25]    (439,149) -- (499,149) ;
\draw [shift={(503,149)}, rotate = 180] [color={rgb, 255:red, 0; green, 0; blue, 0 }  ][line width=2.25]    (17.49,-5.26) .. controls (11.12,-2.23) and (5.29,-0.48) .. (0,0) .. controls (5.29,0.48) and (11.12,2.23) .. (17.49,5.26)   ;
\draw [line width=2.25]    (438,252) -- (498,252) ;
\draw [shift={(502,252)}, rotate = 180] [color={rgb, 255:red, 0; green, 0; blue, 0 }  ][line width=2.25]    (17.49,-5.26) .. controls (11.12,-2.23) and (5.29,-0.48) .. (0,0) .. controls (5.29,0.48) and (11.12,2.23) .. (17.49,5.26)   ;
\draw [line width=2.25]    (437,40) -- (497,40) ;
\draw [shift={(501,40)}, rotate = 180] [color={rgb, 255:red, 0; green, 0; blue, 0 }  ][line width=2.25]    (17.49,-5.26) .. controls (11.12,-2.23) and (5.29,-0.48) .. (0,0) .. controls (5.29,0.48) and (11.12,2.23) .. (17.49,5.26)   ;

\draw (-5,35) node [anchor=north west][inner sep=0.75pt]  [align=left] {{{\huge $\mathbf{X}_{\mathbf{T}}$}}};
\draw (185,30) node [anchor=north west][inner sep=0.75pt]  [align=left] {{ {\Huge $f_T$}}};
\draw (-5,135) node [anchor=north west][inner sep=0.75pt]  [align=left] {{ {\huge $\mathbf{X}_{\mathbf{A}}$}}};
\draw (185,130) node [anchor=north west][inner sep=0.75pt] [align=left] {{ {\Huge $f_A$}}};
\draw (-5,235) node [anchor=north west][inner sep=0.75pt]   {{{\huge $\mathbf{X}_\mathbf{V}$}}};
\draw (185,235) node [anchor=north west][inner sep=0.75pt]  [align=left] {{ {\Huge $f_V$}}};
\draw (400,30) node [anchor=north west][inner sep=0.75pt]   [align=left] {{{\huge $\mathbf{h}_{\mathbf{T}}$}}};
\draw (400,135) node [anchor=north west][inner sep=0.75pt]   [align=left] {{{\huge $\mathbf{h}_{\mathbf{A}}$}}};
\draw (400,235) node [anchor=north west][inner sep=0.75pt]   [align=left] {{{\huge $\mathbf{h}_{\mathbf{V}}$}}};
\draw (543,130) node [anchor=north west][inner sep=0.75pt]   [align=left] {{ {\Huge $f_F$}}};

\end{tikzpicture}

%% file: tex/method.tex
\section{$\mathcal{P}$owMix}
\label{sec:method}

Our proposed multimodal regularization method, \ours, is described here in detail. It consists of five key elements:
\begin{enumerate}
	\item Generating a \emph{varying number of mixed examples}.
	\item \emph{Mixing factor reweighting}, adjusting the contribution of each representation in a mixed example.
	\item \emph{Anisotropic mixing}, \ie, sampling distinct mixing factors for each latent modality space.
	\item \emph{Dynamic mixing}, allowing the combination of a variable number of embeddings from the mini-batch.
	\item \emph{Cross-modal label mixing}, creating a unified multimodal label for each mixed multimodal tuple.
\end{enumerate}
While the first two elements build upon established concepts in the literature (see \autoref{sec:multimix} and \autoref{sec:repr_reweight} respectively), the other three are entirely novel contributions, specifically designed to make \ours perform best in a multimodal setup. In the following, we provide a detailed account of how \ours combines these novel elements in a multi-step algorithm, and then we discuss their motivation and impact. We give pseudo-code for \ours in \autoref{alg:powmix}.


\begin{algorithm}
\small

\SetFuncSty{textsc}
\SetDataSty{emph}
\newcommand{\commentsty}[1]{{\color{ForestGreen}#1}}
\SetCommentSty{commentsty}
\SetKwComment{Comment}{$\triangleright$ }{}

\KwIn{hidden representations $\vH_m \in \real^{B \times d_m}$}
\KwIn{label vector $\vy \in \real^B$}
\KwIn{number $n_O$ of generated mixed examples}
\KwOut{mixed representations $\wt{\vH}_m \in \real^{n_O \times d_m}$}
\KwOut{mixed label vector $\wt{\vy} \in \real^{n_O}$}

$\va_m \gets g_A(\vH_m)$ \Comment*[f]{attention vector} \\
$\valpha_m \sim U(0.5, 2)$ \Comment*[f]{interpolation hyperparameter} \\
$\vLambda_{m} \sim \Dir(\valpha_m)$ \Comment*[f]{interpolation matrix} \\
$\vP \sim U(2, 4)$ \Comment*[f]{masking hyperparameter} \\
$\vM \sim \Bern(\vP / B)$ \Comment*[f]{mask sampling} \\
$\wt{\vLambda}_m \gets \eta_1(\va_m\tran \odot \vM \odot \vLambda_m)$ \Comment*[f]{normalization} \\
$\wt{\vH}_m \gets \wt{\vLambda}_m \vH_m$ \Comment*[f]{representation mixing} \\
$\wt{\vy}_m \gets \wt{\vLambda}_m \vy_m$ \Comment*[f]{label mixing} \\
$\wt{\vy} \gets \frac{1}{M} \sum_m \wt{\vy}_m$ \Comment*[f]{cross modal label mixing}

\caption{\ours multimodal mixing.}
\label{alg:powmix}
\end{algorithm}

\subsection{Algorithm}
\label{sec:powmix_overview}

Given a multimodal example $\vX^{(i)} = (\vX^{(i)}_1, \dots, \vX^{(i)}_M) \in \cX_\cM$, we obtain a hidden representation $\vh^{(i)}_m = f_m(\vX^{(i)}) \in \real^{d_m}$ as the output of encoder $f_m$ for each modality $m \in \cM$. At the mini-batch level, let the matrix $\vH_m \in \real^{B \times d_m}$ hold the hidden representations of all examples in its rows, where $B$ is the mini-batch size. Let also vector $\vy \in \real^B$ denote the labels $y^{(i)} \in \real$ for all examples of the mini-batch.

\paragraph*{Mixing factor reweighting}

First we compute the \emph{mixing factor attention weights} $\va_m \in \real^B$. Specifically, as a form of attention, we use average pooling over the feature dimension followed by ReLU and normalization across modalities:
\begin{equation}
	\va_m = g(\vH_m) \defn
		\frac{\sigma(\vH_m \vone_{d_m} / d_m)}{\sum_{m'} \sigma(\vH_{m'} \vone_{d_{m'}} / d_{m'})},
\label{eq:attention}
\end{equation}
where $\vone_{d_m} \in \real^{d_m}$ is an all-ones vector, $\sigma(\cdot)$ is the ReLU function and  division is performed element-wise. This operation is similar to transformer cross-attention between query $\vone_{d_m}$ and key $\vH_m$ but here normalization is performed across modalities. We call the use of weights $\va_m$ \emph{mixing factor reweighting}. A baseline is to use a uniform $\va = \vone_B / M$ for all modalities.

\paragraph*{Anisotropic mixing}

For each modality, we then sample a different \emph{mixing matrix} $\vLambda_m \in \real^{n_O\times B}$. To do this, we sample $n_O$ distinct $B$-dimensional interpolation coefficient vectors from a Dirichlet distribution $\Dir(\valpha_m)$, with its parameter $\valpha_m \in \real^{n_O}$ drawn from a uniform distribution $U(0.5, 2.0)$, following~\cite{venkataramanan2023Embedding}. Let's assume that the latentrepresentation of each modality is embedded in separate subspaces of an overall multimodal latent representation space. Since we use a different mixing matrix per subspace, we call this approach \emph{anisotropic mixing}. A simpler alternative is to use the same $\vLambda \in \real^{n_O\times B}$ for all modalities.

\paragraph*{Dynamic mixing}

Next, we randomly mask mixing factors to keep only a small number of nonzero elements per row of $\vLambda_m$. This limits the interpolation to just a few examples. Specifically, we sample a binary mask $\vM \sim \Bern(\vP / B) \in \real^{n_O\times B}$ from a Bernoulli distribution and apply it element-wise to the mixing matrix $\vLambda_m$.

The hyperparameter $\vP \in \real^{n_O \times B}$ provides a different value for each element of $\vM$ and dictates the proportion of mini-batch examples to interpolate in the representation space. We find empirically that $\vP \sim U(2, 4)$ works well, corresponding to $2$ to $4$ nonzero elements on average per row of $\vLambda_m$.

Since different examples are interpolated for each generated example, we call this process \emph{dynamic mixing}. Importantly, the same binary mask $\vM$ is used for all modalities $m$ by default, a choice called \emph{mask sharing}. An alternative is to use a different $\vM_m \in \real^{n_O\times B}$ for each modality. The simplest approach would be not to use any masking, \ie setting $\vM = \vone_{n_O \times B}$, which deactivates dynamic mixing. 

\paragraph*{Interpolation operation}

Given the mixing factor attention weights $\va_m \in \real^B$ and the binary mask $\vM \in \real^{n_O\times B}$, we multiply element-wise with the mixing matrix $\vLambda_m \in \real^{n_O\times B}$ and re-normalize over the mini-batch dimension to obtain the \emph{reweighted mixing matrix}
\begin{equation}
	\wt{\vLambda}_m = \eta_1(\va_m\tran \odot \vM \odot \vLambda_m) \in \real^{n_O\times B},
\label{eq:mixing_factor_reweighting_powmix}
\end{equation}
where $\eta_1$ here denotes $\ell_1$-normalization of rows and $\odot$ is Hadamard product (with vectors broadcasting to matrices as needed). In this product, attention weights $\va_m$ are scaling the columns of the mixing matrix, while the binary mask $\vM$ is selecting subsets from each row.

Now, using $\wt{\vLambda}_m$, we generate $n_O$ mixed examples per modality. In particular, given the hidden representations $\vH_m$ and the labels $\vy$, we perform intra-modal interpolation for modality $m$ by
\begin{align}
	\wt{\vH}_m &= \wt{\vLambda}_m \vH_m \in \real^{n_O \times d_m} \label{eq:powmix_h} \\
	\wt{\vy}_m &= \wt{\vLambda}_m \vy \in \real^{n_O}. \label{eq:powmix_y}
\end{align}
This is similar with~\eq{multimix_h}, \eq{multimix_y}, but $\wt{\vLambda}_m$ is reweighted, masked and unique for each modality. Formally, it expresses convex combinations of different input examples in the representation space and their labels, due to nonnegativity of $\wt{\vLambda}_m$ and its rows summing to $1$.

\paragraph*{Cross-modal label mixing}

Finally, we compute a single multimodal label $\wt{\vy} \in \real^{n_O}$ for the mini-batch by averaging $\wt{\vy}_m$ over modalities:
\begin{equation}
	\wt{\vy} = \frac{1}{M} \sum_m \wt{\vy}_m.
\label{eq:cross-modal_label_mixing}
\end{equation}
This step, called \emph{cross-modal label mixing}, is only meaningful when mixing factor reweighting and anisotropic mixing are used, in which case $\wt{\vy}_m$ are different for each modality.


\subsection{Discussion}
\label{sec:method-discussion}

We now discuss the key features of \ours, to better understand the underlying idea of each algorithmic component.

\paragraph*{Varying number of mixed examples}

\ours enables the generation of a variable number $n_O$ of mixed examples. These mixtures lie in the convex hull of the mini-batch in the representation space. 
The value of $n_O$ is a hyperparameter, which is decoupled from the mini-batch size and much larger in practice. 
As such, it generates a plethora of mixtures, leading to more loss terms being calculated per example during model training.
As in MultiMix~\cite{venkataramanan2023Embedding}, it is hypothesized that this provides a better approximation of the expected risk integral.

\paragraph*{Mixing factor reweighting}

Prior works~\cite{transmix, venkataramanan2023Embedding} perform mixing factor reweighting on dense features of image patches, applied to a unimodal task. By contrast, our approach is inherently multimodal, reweighting the mixing factors of each mini-batch example by normalizing across modalities. 
Our intuition is that by reweighting each modality in a multimodal $m$-tuple according to the sum of its features we are able to reduce the impact of uninformative (near zero) unimodal instances in the representation space.  
Unlike prior work~\cite{transmix, venkataramanan2023Embedding}, our approach is also agnostic to the architecture and pooling mechanism of the input modality encoders. We experimentally verify that it improves performance.

\paragraph*{Anisotropic mixing}

Given $m$ hidden tensors $\vH_m \in \real^{B\times d_m}$, \ours samples a separate mixing matrix $\vLambda_m \in \real^{n_O \times B}$ for each modality $m$. This enables modality-specific mixing, that is, the ability of the algorithm to exhibit different mixing strategies across modalities. This property is shown to be critical for \ours to perform well.

\paragraph*{Dynamic mixing and power set}

Considering the formation of the reweighted mixing matrix $\wt{\vLambda}_m$~\eq{mixing_factor_reweighting_powmix}, we track two sources of randomness. The first is due to mixing factors in $\vLambda_m$, sampled from the Dirichlet distribution, which is common in prior mixing methods, \eg~\cite{venkataramanan2023Embedding}.
The second is due to the binary mask $\vM$, sampled from the Bernoulli distribution, which is unique to our method. 
One may interpret the interpolation process as sampling a subset of examples from the \emph{power set} $\cP$ of the mini-batch in the representation space. The subset sampling is then followed by the formation of a convex combination based on the selected representations. 
This is a \emph{dynamic mixing} process in the sense of using a different subset for each generated mixed example. 
The \ours acronym of  the proposed method  alludes to the use of the power set $\cP$.

In practice, the effect of sampling a subset prior to forming a convex combination is that $\wt{\vLambda}_m$ is \emph{sparse}, having a small number of nonzero entries per row (2 to 4 on average according to our default settings). Thus, only few mini-batch examples are interpolated for each generated mixture. 
While it is possible to control the entropy of mixing factors by adjusting the hyperparameter $\valpha_m$ of the Dirichlet distribution in sampling $\vLambda_m$ itself, it is shown experimentally that true sparsity is superior in our multimodal problem. 
As shown in \autoref{fig:powmix_3} and \autoref{fig:powmix_4}, \ours can provide mixing instances between interpolation of \emph{pairs}, as in Manifold Mixup~\cite{verma2019manifold}, and \emph{all} mini-batch examples, as in MultiMix~\cite{venkataramanan2023Embedding}. 
Sharing the binary mask $\vM$ across modalities is empirically shown to be essential for \ours to work well.

\paragraph*{Cross-modal label mixing}

\ours uses mixing factor reweighting and anisotropic mixing, which result in different reweighted mixing matrices $\wt{\vLambda}_m$~\eq{mixing_factor_reweighting_powmix} and thus different mixed labels $\wt{\vy}_m$~\eq{powmix_y} for each modality $m$.
To unify these into a single label vector $\wt{\vy}$, \ours averages the mixed labels $\wt{y}_m$ over modalities~\eq{cross-modal_label_mixing}. This assumes equal contribution from each modality in label generation.
While this assumption may not hold universally, empirical evidence has demonstrated its effectiveness in practice.
The averaging operation is purely multimodal, intertwining label information across different modalities.

%% file: tex/exp-setup.tex
\section{Experimental setup}
\label{sec:exp_setup}

We evaluate \ours and other mixing techniques over three benchmark datasets for MSA and three distinct archetypal multimodal networks. In the following, we provide a detailed description of our experimental setup.


\subsection{Benchmark datasets}

\paragraph*{MOSI}

CMU-MOSI~\cite{zadeh2016multimodal} is an English MSA benchmark dataset consisting of YouTube videos ($\approx 2.5$h), featuring monologues where individuals express opinions, stories and reviews. These videos range from 2-5 minutes in length. CMU-MOSI contains 2199 utterance-level video segments from 93 videos and 89 distinct speakers (41 female and 48 male), with an average segment length of 4.2 sec. Each segment is manually transcribed and annotated with sentiment intensity scores ranging from -3 (strongly negative) to 3 (strongly positive).

\paragraph*{MOSEI}

CMU-MOSEI~\cite{zadeh2018multimodal} is the largest MSA benchmark dataset ($\approx 66$h). Compared to MOSI, it offers a more diverse range of samples, video topics and speakers. MOSEI contains 23,453 manually transcribed and annotated utterance-level video segments from 1000 distinct speakers and covers 250 topics. The average segment length is 7.28 sec, with segmentation based on punctuation from the high-quality manual transcriptions. Each segment is manually annotated in a Likert scale from -3 to 3 as in MOSI.

\paragraph*{SIMS}

The CH-SIMS~\cite{yu2020ch} is a Chinese MSA benchmark dataset, comparable in size to MOSI ($\approx 2.3$h). It consists of 2281 utterance-level monologue video segments from 60 diverse videos, including movies, TV series, and variety shows. Each segment is manually segmented, transcribed, and annotated with sentiment intensity scores ranging from -1 (strongly negative) to 1 (strongly positive). While SIMS provides both multimodal and unimodal annotations, we only leverage the multimodal labels. The average length of each video segment is 3.67 sec.


\subsection{Multimodal features}

Processing raw multimodal video streams is computationally intensive and might also face copyright issues. Therefore, benchmarks in this field typically include a set of extracted features ~\cite{zadeh2016multimodal, zadeh2018multimodal, yu2020ch}. Since feature extraction for emotion and sentiment is a challenge with varied approaches ~\cite{mao-etal-2022-sena}, direct algorithm comparison can be problematic. In our study, we utilize the feature set provided in ~\cite{mao-etal-2022-sena} for fair comparison across benchmarks.

\paragraph*{Text modality}

Semantic word embeddings mainly rely on pretrained language models. Following~\cite{mao-etal-2022-sena}, we use BERT~\cite{devlin2019bert} embeddings adopted from their open-source transformer implementations~\cite{wolf2020transformers}. In particular, we use \texttt{bert-base-uncased} for English and \texttt{bert-base-chinese} for the Chinese language. Eventually, each word is tokenized and represented as a 768-dim word vector.

\paragraph*{Acoustic modality}

The acoustic modality predominantly uses hand-crafted features. MOSI and MOSEI employ the COVAREP~\cite{degottex2014covarep} framework to extract low-level descriptors (LLDs) like pitch and 12 Mel-freq cepstral coefficients (MFCCs), yielding a 74-dim frame-level feature. For SIMS, we use Librosa~\cite{mcfee2015librosa},
resulting in a 33-dim acoustic representation per frame.

\paragraph*{Video modality}

Standard video features for MSA tasks include facial landmarks, eye gaze, and facial action units. For MOSI and MOSEI, 35 facial action units are extracted using Facet \footnote{\url{https://imotions.com/platform}}, focusing on emotion-related movements. In SIMS, the OpenFace2.0 toolkit~\cite{baltrusaitis2018openface} extracts 68 facial landmarks, 17 facial action units, and other features, forming a 709-dim frame-level representation.


\subsection{Evaluation metrics}
\label{sec:eval_metrics}

\paragraph*{Performance metrics}

Aligning with existing literature ~\cite{tsai2019multimodal, hazarika2020misa, yu2021learning, mao-etal-2022-sena}, we evaluate MSA as a regression task using \emph{mean absolute error} (MAE) and \emph{Pearson correlation} (Corr). \emph{Classification accuracy}, denoted as Acc-$k$ for $k$ classes, is also used by mapping regression scores to discrete categories. For binary metrics (Acc-$2$, $F1$), in line with~\cite{tsai2019multimodal, mao-etal-2022-sena}, we exclude neutral (zero-valued) predictions, focusing on positive versus negative values.

\paragraph*{Robustness metrics}

Evaluating MSA model robustness, especially against textual modality dominance, is crucial ~\cite{GKOUMAS2021WhatMakesTheDiff, jin2023weakening}. We assess model robustness under various noisy inputs using the same metrics as for performance.

\begin{table}
\caption{Properties of MSA archetypal models. \emph{FT-BERT}: BERT encoder fine-tuning; \emph{Backbone}: unimodal encoder; \emph{Objectives}: number of objectives in learning recipe. CA: cross-attention; SA: self-attention.}
\label{tab:baseline_models}
\centering
\footnotesize
\setlength{\tabcolsep}{2.2pt}
\begin{tabular}{lcccccc}
\toprule
\Th{Model} & \Th{FT-BERT} & \Th{Backbone} & \Th{Early / Late Fusion} & \Th{Objectives} \\ \midrule
MulT       &              & 1D-CNN        & CA / Concat              & 1               \\
MISA       & \ch          & LSTM          & --- / SA + Concat                & 4               \\
Self-MM    & \ch          & LSTM          & --- / Concat             & 4               \\
\bottomrule
\end{tabular}
\end{table}

\subsection{Competitors and MSA models}

We compare our \ours multimodal regularization method against Manifold MixUp~\cite{verma2019manifold} and state-of-the-art MutiMix~\cite{venkataramanan2023Embedding}, as discussed in \autoref{sec:background}.

All comparisons are performed on three different archetypal MSA architectures, namely MulT~\cite{tsai2019multimodal}, MISA~\cite{hazarika2020misa} and Self-MM~\cite{yu2021learning}. These models have demonstrated strong performance across the datasets we examine~\cite{mao-etal-2022-sena}. As shown in \autoref{tab:baseline_models}, they represent a diverse range of architectural choices, including LSTM, CNNs, and transformers. They also employ various fusion strategies and unique learning recipes, including single-task and multi-task objectives. We reproduce those three models for fair comparison. We also report other models from the literature to provide a holistic performance overview.

\paragraph*{\textbf{LF-DNN}}

The \emph{late fusion deep neural network} (LF-DNN)~\cite{cambria2018benchmarking} learns unimodal features separately for each modality, then concatenates them for multimodal prediction.

\paragraph*{\textbf{TFN}}

The \emph{tensor fusion network} (TFN)~\cite{zadeh2017tensor} employs LSTM for text and averages acoustic and visual features. Latent representations from DNN-processed modalities are concatenated, forming a high-dimensional multimodal space.

\paragraph*{\textbf{MAG-BERT}}

In the MAG-BERT~\cite{rahman2020integrating} model, a multimodal adaptation gate is introduced and integrated with a pretrained BERT backbone to handle multimodal information processing.

\paragraph*{\textbf{MulT}}

The \emph{multimodal transformer} (MulT)~\cite{tsai2019multimodal} employs a 1D-CNN as a unimodal backbone for reducing the dimensionality of input features. MulT uses early fusion through \emph{cross-attention} (CA) blocks. These blocks facilitate interaction and integration of information across different modalities. After this early fusion step, the model processes the combined multimodal streams using \emph{self-attention} (SA) mechanisms. The output of these processes is then concatenated for the final prediction. All mixing operations are integrated prior to concatenation. MulT is optimized based on a single task loss.

\paragraph*{\textbf{MISA}}

The MISA model~\cite{hazarika2020misa} uses LSTM networks to process audio and video modalities. For the text modality, it fine-tunes the BERT encoder. MISA is designed to embed unimodal representations into both a shared multimodal space and distinct unimodal spaces. This design promotes the extraction of common features across modalities while also preserving modality specific features.

In the final stages, MISA decodes all these six representations (from both common and individual spaces) in one branch for reconstruction, while in another branch, it independently merges them using SA to make the final prediction. MISA thus combines a task loss with reconstruction, repelling and attractive objectives. All mixing algorithms are employed before the SA block, treating the six representations as independent modalities.

\paragraph*{\textbf{Self-MM}}

The Self-MM model~\cite{yu2021learning} also relies on LSTM networks to process the audio and visual features. For the text modality, it fine-tunes the BERT encoder. Self-MM implements a pseudolabeling component called \emph{unimodal label generation module} (ULGM), which generates unimodal labels from the multimodal label and the unimodal embeddings. These generated labels then influence the learning process through backpropagation. For the final prediction, Self-MM concatenates the unimodal representations and processes them through a dual linear layer setup. The model combines a task loss with an additional loss for each modality, derived from the pseudolabeling network. All mixing operations are integrated before concatenation.


\subsection{Implementation details}

We employ the M-SENA framework~\cite{mao-etal-2022-sena} for MSA model evaluation, implementing all models in PyTorch~\cite{paszke2019pytorch} and conducting all experiments on a single NVIDIA RTX 3090. We use the Adam~\cite{kingma2014adam} optimizer with early stopping and set hyperparameters per M-SENA's guidelines\footnote{\url{https://github.com/thuiar/MMSA/blob/master/src/MMSA/config/config_regression.json}}. Results for MulT, MISA and Self-MM are reproduced from open-source implementations for fair comparison.

Mixing is integrated before the late fusion stage in training and excluded during inference. For Self-MM model, following official recommendations~\cite{yu2021learning}, we initiate mixing only after the first two epochs for SIMS and MOSEI, and after one epoch for MOSI.
Metrics are averaged over at least five independent runs, while robustness assessments use 15 runs, following~\cite{jing2020self, mao-etal-2022-sena}.

For \ours, we sample the masking probability as $\vP \sim U(2, 4)$ and the interpolation hyperparameter as $\valpha \sim U(0.5, 2.0)$. For MultiMix, we use the default hyperparameters from~\cite{venkataramanan2023Embedding}, while Manifold MixUp performs best with $\alpha = 1.0$. When tuning hyperparameters such as the probability $p_{\text{mix}}$ of applying the mixing algorithm, as well as the number $n_O$ of generated mixed examples for MultiMix and \ours, we employ the following strategy. Initially, $n_O$ is set to 256, and $p_{\text{mix}}$ is optimized. Subsequently, $n_O$ is optimized based on the optimal $p_{\text{mix}}$ value. The process may be repeated if the results are not satisfying. For Manifold MixUp, because of the small batch size, we process batches twice the baseline size, split each batch in half and perform mixing between the two halves.

%% file: tex/exp-results.tex
\section{Experimental results}
\label{sec:experiments}

We evaluate \ours against competitors over a diverse set of multimodal networks across different MSA benchmark datasets. All latent feature regularizers are applied before the late fusion part of each architecture.

\begin{table*}
\caption{\emph{State of the art comparisons}. M.MixUp: Manifold MixUp. $^\dag$: results reported in~\cite{mao-etal-2022-sena}; all other results are reproduced. $\uparrow/\downarrow$: higher/lower is better. Red: worse than the baseline; bold: best for each MSA model.}
\label{tab:mosi_mosei_complete}
\centering
\footnotesize
\setlength{\tabcolsep}{3pt}
\begin{tabular}{lccccccccccccccccccccccccc}
\toprule
\mr{2}{\Th{Model}} & \mc{6}{MOSI} & \mc{6}{MOSEI} & \mc{4}{SIMS} \\ \cmidrule(lr){2-7} \cmidrule(lr){8-13} \cmidrule(lr){14-17}
& \Th{Acc2}$\uparrow$ & F1$\uparrow$ & MAE$\downarrow$ & \Th{Corr}$\uparrow$ & \Th{Acc5}$\uparrow$ &  Acc7$\uparrow$ & \Th{Acc2}$\uparrow$ & F1$\uparrow$ & MAE$\downarrow$ & \Th{Corr}$\uparrow$ & \Th{Acc5}$\uparrow$ &  Acc7$\uparrow$ & \Th{Acc2}$\uparrow$ & F1$\uparrow$ & MAE$\downarrow$ & \Th{Corr}$\uparrow$ \\

\midrule
LF-DNN$^\dag$    & 79.39 & 79.45 & 0.945 & 0.675 & - & -                   & 82.78 & 82.38 & 0.558 & 0.731 & - & -      & 76.68 & 76.48 & 0.446  & 0.567           \\
TFN$^\dag$    & 78.02 & 78.09 & 0.971 & 0.652 & - & -                   & 82.23 & 81.47 & 0.573 & 0.718 & - & - & 77.07 & 76.94 & 0.437  & 0.582       \\
MAG-BERT$^\dag$    & 83.41 & 83.47 & 0.761 & 0.772 & - & -                   & 84.87 & 84.85 & 0.539 & 0.764 & - & - & 74.44 & 71.75 & 0.492  & 0.399           \\
MulT$^\dag$    & 80.21 & 80.22 & 0.912 & 0.695 & - & -                   & 84.63 & 84.52 & 0.559 & 0.733 & - & - & 78.56 & 79.66 & 0.453  & 0.564    \\
MISA$^\dag$    & 82.96 & 82.98 & 0.761     & 0.772     & - & -           & 84.63 & 84.52 & 0.559 & 0.733 & - & -  & 76.54 & 76.59 & 0.447  & 0.563   \\
Self-MM$^\dag$ & 84.30 & 84.31 & 0.720     & 0.793     & - & -           & 84.06 & 84.12 & 0.531 & 0.766 & - & - & 80.04 & 80.44 & 0.425  & 0.595   \\

\midrule
MulT    & 80.26 & 80.32 & 0.927 & 0.689 & 40.10 & 34.71                   & 84.07 & 83.93 & 0.564 & 0.731 & 53.97 & 52.56 &77.77&77.99&0.442&0.584  \\
+ M.MixUp    & 80.41 & 80.36 & \rd{0.928} & \rd{0.686} & \rd{39.14} & \rd{34.26}                   & \rd{84.02} & \rd{83.92} & 0.563 & \rd{0.729} & 54.19 & \rd{52.50} & 78.09 & \rd{77.95} & \rd{0.445} & \rd{0.576}           \\
+ MultiMix    & 80.46 & 80.49 & 0.911 & \rd{0.688} & \rd{39.33} & 34.96                   & 84.08 & 84.01 & 0.563 & 0.733 & 53.99 & \rd{52.39}  & 78.09 & \rd{77.87} & \rd{0.445} & \rd{0.575}          \\
\rowcolor{Gray3}
+ \ours    & \tb{81.01} & \tb{80.99} & \tb{0.904} & \tb{0.696} & \tb{40.65} & \tb{35.00}                   & \tb{84.44} & \tb{84.38} & \tb{0.559} & \tb{0.738} & \tb{54.26} & \tb{52.75} & \tb{79.04} & \tb{78.51} & \tb{0.437} &  \tb{0.595}          \\

\midrule
MISA    & 82.93 & 82.95 & 0.772     & 0.774     & 47.55 & 42.10           & 84.51 & 84.47 & 0.549 & 0.759 & 53.57 & 51.96   &76.59&76.20&0.457&0.550           \\
+ M.MixUp    & 83.08 & 83.12 & \rd{0.783}     & \rd{0.770}     & \rd{46.94} & 42.10           & \rd{84.50} & \rd{84.32} & \rd{0.551} & \rd{0.755} & 53.61 & 52.10   &\rd{75.60} & \rd{75.47} &\rd{0.460} & \rd{0.549}          \\
+ MultiMix    & \rd{82.82} & \rd{82.86} & \rd{0.780}  & 0.778 & \rd{47.06} & \rd{41.80}           & 84.55 & 84.47 & \rd{0.551} & \rd{0.757} & 53.94 & 52.30 &76.67&\rd{76.15}&0.455&\rd{0.547}         \\
\rowcolor{Gray3}
+ \ours    & \tb{83.49} & \tb{83.50} & \tb{0.761}     & \tb{0.780}     & \tb{48.02} & \tb{42.65}           & \tb{84.97} & \tb{84.86} & \tb{0.543} & \tb{0.762} & \tb{54.52} & \tb{53.00} & \tb{77.35} & \tb{76.97} & \tb{0.441} & \tb{0.569}         \\

\midrule
Self-MM & 84.22 & 84.23 & 0.724     & 0.791     & 52.22 & 45.64           & 84.26 & 84.24 & 0.532 & 0.765 & 55.52 & 53.85 &78.16&78.15&0.417&0.592         \\
+ M.MixUp & 84.38 & 84.37 & 0.722     & 0.792     & 53.50 & 46.33           & \rd{84.24} & \rd{84.23} & 0.532 & 0.765 & 55.57 & \rd{53.82} & 78.56 &  78.58 &  0.414 & 0.594         \\
+ MultiMix & 84.35 & 84.38 & 0.723  & 0.792 & 52.45 & 45.89           & \rd{84.17} & \rd{84.16} & \rd{0.547} & \rd{0.751} & \rd{54.53} & \rd{52.84} & \rd{77.62} & \rd{77.77} & \rd{0.426} & \rd{0.576}         \\
\rowcolor{Gray3}
+ \ours & \tb{84.76} & \tb{84.78} & \tb{0.712}     & \tb{0.795}     & \tb{53.86} & \tb{46.88}           & \tb{85.11} & \tb{85.10} & \tb{0.528} & \tb{0.770} & \tb{55.87} & \tb{54.25} & \tb{79.02} & \tb{78.94} & \tb{0.412} & \tb{0.599}   \\
\bottomrule
\end{tabular}
\end{table*}

\subsection{Comparison with the state of the art}
\label{sec:main_results}

To ensure a fair comparison between different mixing methods and consistency across all evaluation metrics, we reproduce MulT, MISA and Self-MM baseline models. Additionally, we present results from established frameworks, which are generally comparable with our reproduced results. We primarily compare our results to the reproduced ones.

\autoref{tab:mosi_mosei_complete} evaluates \ours against the state of the art. These results clearly show that integrating \ours leads to consistent performance improvements across all metrics and datasets over the reproduced baseline models. This result clearly illustrates the width of applicability and consistency of the proposed method across various architectures, fusion schemes and learning recipes. Notably, in the vast majority of the examined metrics ($72\%$), Manifold MixUp and MultiMix fail to improve or even harm performance compared to the baseline. By contrast, all models are benefited by multimodal regularization across all setups. Next, we take a closer look at individual datasets.

\paragraph*{MOSI}
By using \ours, Self-MM outperforms all examined models, both reproduced and original. For MulT, improves by $0.75$ Acc-$2$ and $0.023$ MAE.

\paragraph*{MOSEI}
By using \ours, MISA improves by $0.95$ Acc-$5$ and $1.04$ Acc-$7$. Self-MM improves by $0.86$ for binary metrics, outperforming its original results by $1.0$ on average and even outperforming MISA, which has better baseline performance on these metrics.

\paragraph*{SIMS}
\ours significantly boosts all models. By using \ours, MulT outperforms the stronger baseline Self-MM model across three metrics, which clearly confirms the benefits of regularization in multimodal architectures.

%% file: tex/exp-ablation.tex
\subsection{Ablation study}

To investigate the effect of each hyperparameter and algorithmic component in \ours, we conduct extensive experiments on MOSEI (largest benchmark) with Self-MM (best performing model).

\begin{figure}
\centering
\begin{subfigure}[b]{0.9\columnwidth}
	\centering
	\begin{tikzpicture}
	\begin{axis}[
		font=\scriptsize,
		bar width=0.25cm,
		xbar, 
		width=\columnwidth, 
		xmin=84.7, 
		xmax=85.2,
		xlabel={Acc-$2$ ($\uparrow$)}, 
		ylabel={}, 
		symbolic y coords={$U{(2,16)}$, $U{(2,8)}$, $U{(6,8)}$, $U{(4,6)}$, $U{(2,4)}$}, 
		ytick=data, 
		nodes near coords, 
		nodes near coords style={color=Gray0},
		xmajorgrids=true,
		height=4cm,
	]

	\addplot [draw=CB91_Blue,
	pattern color=CB91_Blue, pattern=north east lines
	]
	coordinates {
		(84.93,$U{(2,16)}$)
		(85.07,$U{(2,8)}$)
		(84.87,$U{(6,8)}$)
		(85.08,$U{(4,6)}$)
		(85.11,$U{(2,4)}$)
	};

	\end{axis}
	\end{tikzpicture}
	\label{fig:acc2_interpolated_examples}
\end{subfigure}
\vspace{5mm}
\begin{subfigure}[b]{0.9\columnwidth}
	\centering
	\begin{tikzpicture}
	\begin{axis}[
		font=\scriptsize,
		bar width=0.25cm,
		xbar, 
		width=\columnwidth, 
		height=4cm, 
		xmin=52.8, 
		xmax=53.25,
		xlabel={MAE ($\times 100$) ($\downarrow$)}, 
		ylabel={}, 
		symbolic y coords={$U{(2,16)}$, $U{(2,8)}$, $U{(6,8)}$, $U{(4,6)}$, $U{(2,4)}$}, 
		ytick=data, 
		nodes near coords, 
		nodes near coords style={color=Gray0},
		xmajorgrids=true,
	]

	\addplot [
	color=CB91_Purple,
	postaction={pattern=north west lines, pattern color=CB91_Purple}
	]
	coordinates {
		(53.10,$U{(2,16)}$)
		(52.91,$U{(2,8)}$)
		(53.15,$U{(6,8)}$)
		(52.95,$U{(4,6)}$)
		(52.86,$U{(2,4)}$)
	};

	\end{axis}
	\end{tikzpicture}
	\label{fig:mae_interpolated_examples}
\end{subfigure}
\caption{The effect of \emph{subset sampling} in \ours, as controlled by the uniform distribution $U(a, b)$ used to sample the hyperparameter $\vP \in \real^{n_O \times B}$ of the Bernoulli distribution from which we sample the binary mask $\vM \in \real^{n_O \times B}$. Using Self-MM model on MOSEI. $\uparrow/\downarrow$: higher/lower is better.}
\label{fig:n_I_barplots}
\end{figure}
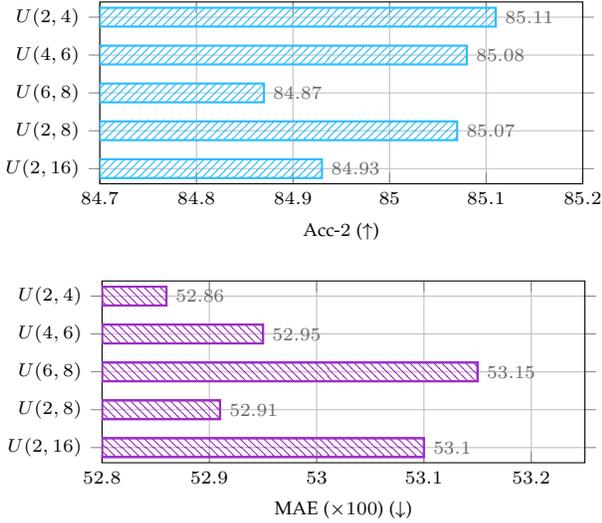

\subsubsection{Subset sampling}
\label{sec:n_I_results}

In this experiment, we perform an ablation on subset sampling from the mini-batch prior to forming convex combinations in \ours. This is controlled by a uniform distribution $U(a, b)$ used to sample the hyperparameter $\vP \in \real^{n_O \times B}$ of the Bernoulli distribution from which we sample the binary mask $\vM \in \real^{n_O \times B}$. This uniform distribution means that there are from $a$ to $b$ nonzero entries on average in each row of the mask $\vM$, thus also each row of the reweighted mixing matrix $\hat{\vLambda}_m$. In turn, this means that we are interpolating from $a$ to $b$ mini-batch examples on average.

\autoref{fig:n_I_barplots} shows the results for a variety of choices for $a, b$. According to both metrics, the intervals are ranked by decreasing performance as $U(2,4)$, $U(2, 8)$, $U(4, 6)$, $U(2, 16)$, and $U(6,8)$. This highlights the importance of sampling a small subset of mini-batch examples (both $a$ and $b$ being small) and the sparsity of $\hat{\vLambda}_m$, justifying our dynamic mixing process. Notably, even the least effective choice, $U(6,8)$, still outperforms the baseline Self-MM model. We choose $U(2, 4)$ by default in \ours based on these results.


\begin{figure}[]
\centering
\begin{subfigure}[h]{0.9\columnwidth}
\centering
\begin{tikzpicture}
\begin{axis}[
	font=\scriptsize,
	xlabel={$n_O$},
	ylabel={Acc-$2$ ($\uparrow$)},
	xmin=12, xmax=2560,
	xtick={16, 32, 64, 128, 256, 512, 1024, 2048},
	xmode=log,
	log basis x={2},
	ytick={84.2, 84.4, 84.6, 84.8, 85, 85.2, 85.4},
	ytick distance=0.25, 
	legend pos=south east, 
	height=.5\columnwidth,
	ymajorgrids=true,
]


\addplot[color=CB91_Violet] coordinates {
(16, 84.26) 
(2048, 84.26) 
};
\addlegendentry{Self-MM}

\addplot[color=my_Green, mark=*] coordinates {
	(16, 84.31)
	(32, 84.56)
	(64, 84.66)
	(128, 84.90)
	(256, 85.11)
	(512, 85.05)
	(1024, 84.89)
	(2048, 84.78)
};
\addlegendentry{Self-MM + \ours}



\end{axis}
\end{tikzpicture}
\label{fig:selfmm_mosei_ablation_acc2}
\end{subfigure}
\hfill
\begin{subfigure}[h]{0.9\columnwidth}
\centering
\begin{tikzpicture}
\begin{axis}[
	font=\scriptsize,
	xlabel={$n_O$},
	xmin=12, xmax=2560,
	xtick={16, 32, 64, 128, 256, 512, 1024, 2048},
	xmode=log,
	log basis x={2},
	ylabel={MAE ($\downarrow$)},
	ytick={52.8, 52.9, 53.0, 53.1, 53.2},
	ytick distance=0.1, 
	legend pos=north east, 
	height=.5\columnwidth,
	ymajorgrids=true,
]

\addplot[color=CB91_Violet, thick] coordinates {
(16, 53.20) 
(2048, 53.20) 
};


\addplot[color=my_Green, mark=*, thick]
coordinates {
	(16, 0.5322*100)
	(32, 0.5308*100)
	(64, 0.5302*100)
	(128, 0.5291*100)
	(256, 0.5286*100)
	(512, 0.5287*100)
	(1024, 0.5301*100)
	(2048, 0.5299*100)
};



\end{axis}
\end{tikzpicture}
\label{fig:selfmm_mosei_ablation_mae}
\end{subfigure}
\caption{Impact of the number $n_O$ of \emph{generated mixed examples} on MOSEI. $\uparrow/\downarrow$: higher/lower is better.}
\label{fig:nO_ablation}
\end{figure}
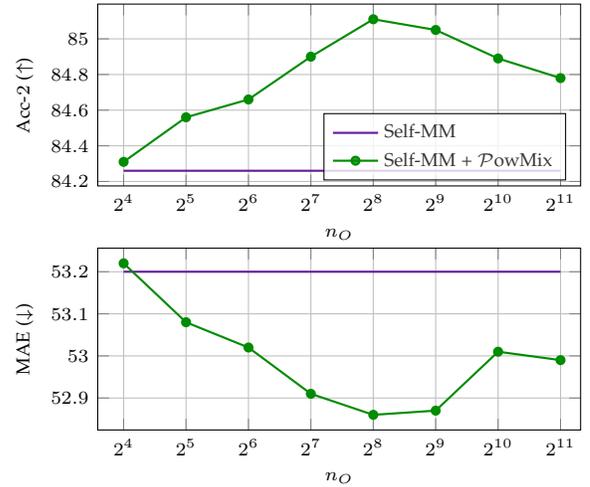

\subsubsection{Number of generated mixed examples}
\label{sec:n_O_results}

\autoref{fig:nO_ablation} shows the effect of the number $n_O$ of generated mixed examples. We observe that a smaller number of generated mixed examples, such as $2^4$ or $2^5$, does not significantly benefit the model, whereas a very large number, like $2^{11}$, seems suboptimal. The best-performing values are found among $\{2^7, 2^8, 2^9\}$, with $2^8$ performing best. We underline that \ours is effective over a very wide range of $n_O$ values, outperforming the baseline Self-MM model for all values tested.

\begin{table}
\caption{Effect of \emph{algorithmic components} of \ours: anisotropic mixing (Aniso.), mixing factor reweighting (Reweight), cross-modal mask sharing (M. Share), and dynamic mixing (D.Mix). Using Self-MM model on MOSEI. $\uparrow/\downarrow$: higher/lower is better.}
\label{tab:ablation_mixing_config}
\centering
\footnotesize
\setlength{\tabcolsep}{3pt}
\begin{tabular}{ccccccccc}
\toprule
\Th{Aniso.} & \Th{Reweight} & \Th{M.Share} & \Th{D.Mix} & \Th{Acc2}$\uparrow$  & \Th{Acc5}$\uparrow$ & MAE$\downarrow$ \\
\midrule \rowcolor{Gray3}
\ch & \ch & \ch & \ch & \tb{85.11} & \tb{55.87} & \tb{0.528}  \\
    & \ch & \ch & \ch & 84.90 & 55.60 & 0.529  \\
\ch &     & \ch & \ch & 85.00 & 55.77 & \tb{0.528}  \\
\ch & \ch &     & \ch & 84.99 & 55.30 & 0.531  \\
\ch & \ch &  &  & 84.62 & 55.21 & 0.535  \\
\bottomrule
\end{tabular}
\end{table}

\subsubsection{Algorithmic components}

By turning four critical algorithmic components on/off during training, we study their effect on \ours. In paricular, we examine anisotropic mixing, mixing factor reweighting, cross-modal mask sharing and dynamic mixing. It is important to note that cross-modal label mixing~\eq{cross-modal_label_mixing} is linked with other components. When turning off mixing factor reweighting and anisotropic mixing, we also turn off cross-modal label mixing, since in this case $\wt{\vy}_m$~\eq{powmix_y} are the same across modalities. When turning dynamic mixing off, i.e., use unit mask in~\eqref{eq:mixing_factor_reweighting_powmix}, we assume that the mask sharing feature is also deactivated.

\autoref{tab:ablation_mixing_config} shows that dynamic mixing, \ie, using sparse binary masks, is the most important component of \ours. Morever, mask sharing, \ie, masking the same examples across modalities, is also a crucial component.
Therefore, we use a sparse shared cross-modal mask in all our experiments.
Anisotropic mixing is also essential, as its absence lowers performance.
The benefit of reweighting is also clear, supporting findings from other studies~\cite{transmix}.
These findings underscore the synergetic impact of the features of \ours; removing any of them leads to a drop in performance.

%% file: tex/exp-analysis.tex
\subsection{Analysis}

To better understand how \ours works and affects the models, we conduct analysis experiments on MOSEI.

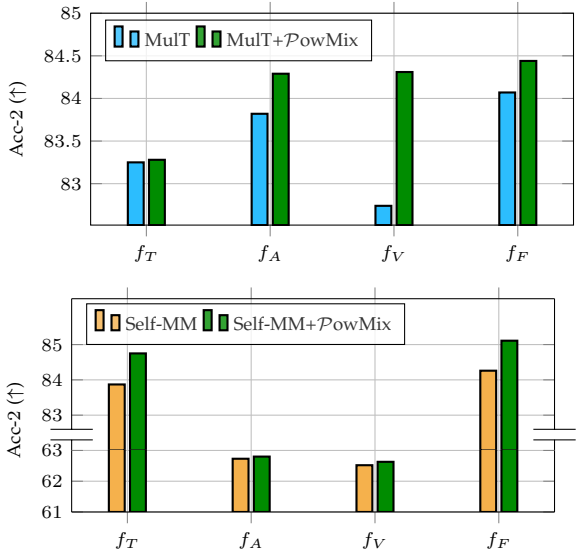
\begin{figure}
\begin{subfigure}{0.9\columnwidth}
\centering
\begin{tikzpicture}
\begin{axis}[
	font=\scriptsize,
	ybar,
	ymax=85,
	enlarge x limits=0.15, 
	ylabel={Acc-$2$ ($\uparrow$)},
	symbolic x coords={$f_T$, $f_A$, $f_V$, $f_F$},
	xtick={$f_T$, $f_A$, $f_V$, $f_F$}, 
	xticklabels={$f_T$, $f_A$, $f_V$, $f_F$}, 
	height=0.55\columnwidth,
	ymajorgrids=true,
	bar width=6pt,
	legend columns=-1,
	legend pos=north west,
]


\addplot+[draw=black, fill=CB91_Blue, bar shift=-4pt] coordinates {($f_T$,83.25) ($f_A$,83.82) ($f_V$,82.74) ($f_F$,84.07)};
\addlegendentry{MulT}

\addplot+[draw=black, fill=my_Green, bar shift=4pt] coordinates {($f_T$,83.28) ($f_A$,84.29) ($f_V$,84.31) ($f_F$,84.44)};
\addlegendentry{MulT+\ours}

\end{axis}
\end{tikzpicture}
\end{subfigure}

\begin{subfigure}[h]{0.9\columnwidth}
\centering
\begin{tikzpicture}
\begin{groupplot}[
	group style={
		group name=my fancy plots,
		group size=1 by 2,
		xticklabels at=edge bottom,
		vertical sep=0pt,
	},
	ybar,
	enlarge x limits=0.15, 
	symbolic x coords={$f_T$, $f_A$, $f_V$, $f_F$},
	xtick={$f_T$, $f_A$, $f_V$, $f_F$}, 
	xticklabels={$f_T$, $f_A$, $f_V$, $f_F$}, 
	ymajorgrids=true,
	legend columns=-1,
	legend pos=north west,
]


\nextgroupplot[
	ymin=82, ymax=86.3,
	ytick={83, ..., 85.3},
	axis x line=top,
	axis y discontinuity=parallel,
	height=0.45\columnwidth,
	font=\scriptsize,
	bar width=6pt,
	ylabel={Acc-$2$ ($\uparrow$)},
    ylabel style={at={(ticklabel cs:0.2)}}
]

\addplot+[draw=black, fill=CB91_Amber, bar shift=-4pt] coordinates {($f_T$,83.87) ($f_F$,84.26)};
\addlegendentry{Self-MM}

\addplot+[draw=black, fill=my_Green, bar shift=4pt] coordinates {($f_T$,84.75) ($f_F$,85.11)};
\addlegendentry{Self-MM+\ours}

\nextgroupplot[
	ymin=61, ymax=63,
	ytick={61, ..., 63},
	axis x line=bottom,
	height=0.30\columnwidth,
	font=\scriptsize,
	bar width=6pt,
]

\addplot+[draw=black, fill=CB91_Amber, bar shift=-4pt] coordinates {($f_T$,83.87) ($f_A$,62.73) ($f_V$,62.52) ($f_F$,84.26)};

\addplot+[draw=black, fill=my_Green, bar shift=4pt] coordinates {($f_T$,84.75) ($f_A$,62.80) ($f_V$,62.63) ($f_F$,85.11)};

\end{groupplot}
\end{tikzpicture}
\end{subfigure}
\caption{\emph{Unimodal evaluation} analysis on MOSEI. Effect of \ours on performance of encoders $f_m$ of individual modality $m \in \{T, A, V\}$ and fusion network $f_F$ for different MSA architectures. $f_m$ evaluated by training a linear head on each modality representations ($\vh_m$), while keeping $f_m$ frozen. $f_F$ evaluated based on \autoref{tab:mosi_mosei_complete}. $T$: text; $A$: acoustic; $V$: video. MulT: early fusion; Self-MM: late fusion. $\uparrow/\downarrow$: higher/lower is better.}
\label{fig:stream_lin_eval}
\end{figure}

\subsubsection{Unimodal evaluation}

In this experiment, we explore how \ours influences each modality before fusion, as well as fusion itself. Referring to \autoref{fig:abstract_fusion}, we denote by $f_m$ each input modality encoder with $m \in \{T, A, V\}$ ($T$: text, $A$: acoustic, $V$: video) and by $f_F$ the fusion network. We train the model with and without \ours. After training, we assess the performance of each modality by training a linear head on the output of each modality representations ($\vh_m$), while keeping $f_m$ frozen. The fusion network $f_F$ is evaluated based on \autoref{tab:mosi_mosei_complete}. We repeat each experiment three times and report the average performance.

The results are shown in \autoref{fig:stream_lin_eval}. All modalities perform well in the MulT architecture ($f_m$), which uses early fusion. Applying \ours improves $f_A$ by $0.47$ Acc-2 and $f_V$ by $1.57$, which is significant, while $f_T$ shows minimal improvement ($0.03$). The fusion network $f_F$ improves notably by $0.37$ Acc-2, though this is less than the per-modality improvement.

By contrast, the Self-MM architecture, which uses late fusion, shows different trends. Here, the acoustic and visual modalities perform substantially worse ($\approx 63$ Acc-2) than the dominant text modality ($83.87$). With \ours, the audio and visual streams show minor improvements ($\approx 0.07$), but the text modality $f_T$ improves significantly by $0.88$ Acc-2. Interestingly, this gain is directly transferred to the fusion network $f_F$, improving it by $0.85$ Acc-2.

These findings are mutually informative. We interpret the results based on architectural differences and unimodal feature analysis from~\cite{du2023onunimodal}. In MulT, the substantial per-modality improvements do not translate to similar multimodal predictive gains. Due to its early fusion approach, the gain of each modality is inherently multimodal, limiting the margin for additional gain at the final fusion stage. Conversely, in Self-MM, the significant boost in the text stream sufficiently enhances the performance of the final fusion layer, which aligns with the concept that unimodal improvements directly benefit the fusion process.

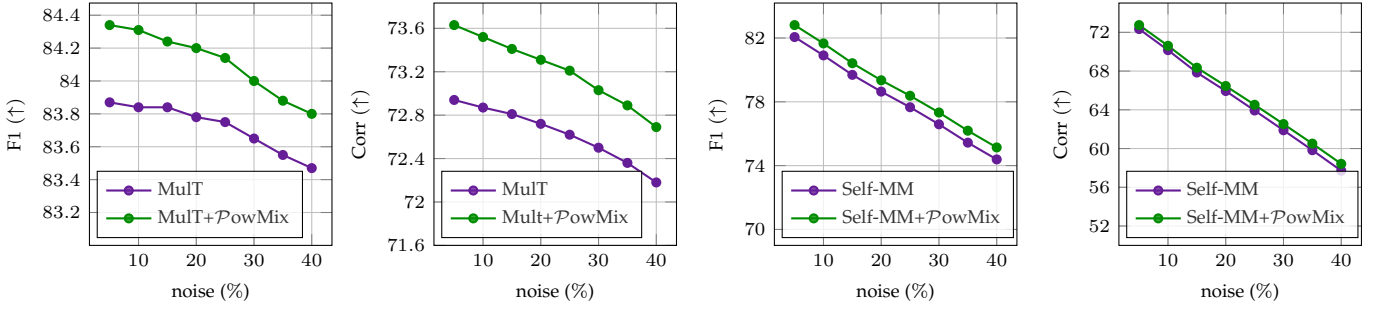
\begin{figure*}
\centering
\begin{subfigure}{0.24\textwidth}
	\centering
	\begin{tikzpicture}
		\begin{axis}[
			font=\scriptsize,
			xlabel={noise ($\%$)},
			ylabel={F$1$ $(\uparrow)$},
			ymin=83.0,
			ytick={83.2, 83.4, ..., 84.4},
			legend pos=south west,
			width=1.1\linewidth,
			height=1.1\linewidth,
		]

		\addplot[color=CB91_Violet, mark=*]
		coordinates {
			(5, 83.87)
			(10, 83.84)
			(15, 83.84)
			(20, 83.78)
			(25, 83.75)
			(30, 83.65)
			(35, 83.55)
			(40, 83.47)
		};
		\addlegendentry{MulT}
		\addplot[color=my_Green, mark=*]
		coordinates {
			(5, 84.34)
			(10, 84.31)
			(15, 84.24)
			(20, 84.20)
			(25, 84.14)
			(30, 84.00)
			(35, 83.88)
			(40, 83.80)
		};
		\addlegendentry{MulT+\ours}
		\end{axis}
	\end{tikzpicture}
\end{subfigure}%
\hfill
\begin{subfigure}{0.24\textwidth}
	\centering
	\begin{tikzpicture}
		\begin{axis}[
			font=\scriptsize,
			xlabel={noise ($\%$)},
			ylabel={Corr $(\uparrow)$},
			ymin=71.6,
			ytick={71.6, 72, ..., 73.6},
			legend pos=south west,
			width=1.1\linewidth,
			height=1.1\linewidth,
		]

		\addplot[color=CB91_Violet, mark=*]
		coordinates {
			(5, 72.94)
			(10, 72.87)
			(15, 72.81)
			(20, 72.72)
			(25, 72.62)
			(30, 72.50)
			(35, 72.36)
			(40, 72.18)
		};
		\addlegendentry{MulT}
		\addplot[color=my_Green, mark=*]
		coordinates {
			(5, 73.63)
			(10, 73.52)
			(15, 73.41)
			(20, 73.31)
			(25, 73.21)
			(30, 73.03)
			(35, 72.89)
			(40, 72.69)
		};
		\addlegendentry{Mult+\ours}
		\end{axis}
	\end{tikzpicture}
\end{subfigure}
\hfill
\begin{subfigure}{0.24\textwidth}
	\centering
	\begin{tikzpicture}
		\begin{axis}[
			font=\scriptsize,
			xlabel={noise ($\%$)},
			ylabel={F$1$ $(\uparrow)$},
			ymin=69,
			ytick={70, 74, ..., 84}, 
			legend pos=south west,
			width=1.1\linewidth,
			height=1.1\linewidth,
		]

		\addplot[color=CB91_Violet, mark=*]
		coordinates {
			(5, 82.06)
			(10, 80.91)
			(15, 79.69)
			(20, 78.64)
			(25, 77.66)
			(30, 76.59)
			(35, 75.44)
			(40, 74.39)
		};
		\addlegendentry{Self-MM}
		\addplot[color=my_Green, mark=*]
		coordinates {
			(5, 82.81)
			(10, 81.66)
			(15, 80.42)
			(20, 79.35)
			(25, 78.38)
			(30, 77.33)
			(35, 76.19)
			(40, 75.14)
		};
		\addlegendentry{Self-MM+\ours}
		\end{axis}
	\end{tikzpicture}
\end{subfigure}%
\hfill
\begin{subfigure}{0.24\textwidth}
	\centering
	\begin{tikzpicture}
		\begin{axis}[
			font=\scriptsize,
			xlabel={noise ($\%$)},
			ylabel={Corr $(\uparrow)$},
			ymin=50,
			ytick={52, 56, ..., 72},
			legend pos=south west,
			width=1.1\linewidth,
			height=1.1\linewidth,
		]

		\addplot[color=CB91_Violet, mark=*]
		coordinates {
			(5, 72.34)
			(10, 70.16)
			(15, 67.85)
			(20, 65.94)
			(25, 63.93)
			(30, 61.89)
			(35, 59.83)
			(40, 57.71)
		};
		\addlegendentry{Self-MM}
		\addplot[color=my_Green, mark=*]
		coordinates {
			(5, 72.75)
			(10, 70.61)
			(15, 68.35)
			(20, 66.46)
			(25, 64.52)
			(30, 62.53)
			(35, 60.51)
			(40, 58.42)
		};
		\addlegendentry{Self-MM+\ours}
		\end{axis}
	\end{tikzpicture}
\end{subfigure}
\caption{\emph{Robustness to noise} analysis on MOSEI. Input frames randomly dropped with probability ranging from 5\% to 40\%. Average metrics reported over dropping being aligned across modalities and independent. $\uparrow/\downarrow$: higher/lower is better.}
\label{fig:rob_mult_selfmm}
\end{figure*}


\subsubsection{Robustness to noise}
\label{sec:robustness}

Next, we investigate the impact of \ours on model robustness. We train MulT and Self-MM models with and without \ours using clean data and then evaluate them in noisy conditions. In particular, we randomly drop input frames from each modality (temporal drop) with a probability $p$ determining the noise intensity and ranging from $5\%$ to $40\%$. This drop occurs either in a correlated fashion (simultaneously across all modalities) or independently. We average the results over the two noise types.

\autoref{fig:rob_mult_selfmm} shows that both MulT and Self-MM are impacted by noise. Interestingly, integrating \ours does not significantly alter the effect of noise, as indicated by the slope of the curves. Models trained with \ours maintain their gain over baselines in noisy conditions. Notably, MulT exhibits greater noise robustness than Self-MM, as indicated by a smaller drop of performance.

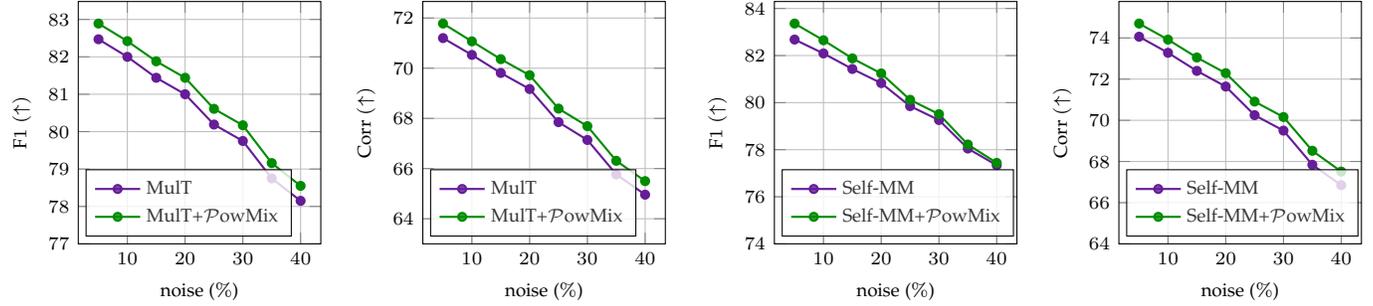
\begin{figure*}
\centering
\begin{subfigure}{0.24\textwidth}
	\centering
	\begin{tikzpicture}
		\begin{axis}[
			font=\scriptsize,
			xlabel={noise ($\%$)},
			ylabel={F$1$ $(\uparrow)$},
			ymin=77,
			ytick={77, 78, ..., 83},
			legend pos=south west,
			width=1.1\linewidth,
			height=1.1\linewidth,
		]

		\addplot[color=CB91_Violet, mark=*]
		coordinates {
			(5, 82.47)
			(10, 82.00)
			(15, 81.44)
			(20, 81.00)
			(25, 80.19)
			(30, 79.75)
			(35, 78.75)
			(40, 78.15)
		};
		\addlegendentry{MulT}
		\addplot[color=my_Green, mark=*]
		coordinates {
			(5, 82.89)
			(10, 82.42)
			(15, 81.88)
			(20, 81.44)
			(25, 80.61)
			(30, 80.17)
			(35, 79.16)
			(40, 78.55)
		};
		\addlegendentry{MulT+\ours}
		\end{axis}
	\end{tikzpicture}
\end{subfigure}%
\hfill
\begin{subfigure}{0.24\textwidth}
	\centering
	\begin{tikzpicture}
		\begin{axis}[
			font=\scriptsize,
			xlabel={noise ($\%$)},
			ylabel={Corr $(\uparrow)$},
			ymin=63,
			ytick={64, 66, ..., 72},
			legend pos=south west,
			width=1.1\linewidth,
			height=1.1\linewidth,
		]

		\addplot[color=CB91_Violet, mark=*]
		coordinates {
			(5, 71.20)
			(10, 70.53)
			(15, 69.81)
			(20, 69.17)
			(25, 67.85)
			(30, 67.14)
			(35, 65.77)
			(40, 64.96)
		};
		\addlegendentry{MulT}
		\addplot[color=my_Green, mark=*]
		coordinates {
			(5, 71.78)
			(10, 71.07)
			(15, 70.36)
			(20, 69.72)
			(25, 68.39)
			(30, 67.69)
			(35, 66.31)
			(40, 65.50)
		};
		\addlegendentry{MulT+\ours}
		\end{axis}
	\end{tikzpicture}
\end{subfigure}
\hfill
\begin{subfigure}{0.24\textwidth}
	\centering
	\begin{tikzpicture}
		\begin{axis}[
			font=\scriptsize,
			xlabel={noise ($\%$)},
			ylabel={F$1$ $(\uparrow)$},
			ymin=74,
			ytick={74, 76, ..., 84}, 
			legend pos=south west,
			width=1.1\linewidth,
			height=1.1\linewidth,
		]

		\addplot[color=CB91_Violet, mark=*]
		coordinates {
			(5, 82.68)
			(10, 82.09)
			(15, 81.43)
			(20, 80.83)
			(25, 79.85)
			(30, 79.26)
			(35, 78.05)
			(40, 77.34)
		};
		\addlegendentry{Self-MM}
		\addplot[color=my_Green, mark=*]
		coordinates {
			(5, 83.36)
			(10, 82.65)
			(15, 81.88)
			(20, 81.24)
			(25, 80.12)
			(30, 79.51)
			(35, 78.22)
			(40, 77.44)
		};
		\addlegendentry{Self-MM+\ours}
		\end{axis}
	\end{tikzpicture}
\end{subfigure}%
\hfill
\begin{subfigure}{0.24\textwidth}
	\centering
	\begin{tikzpicture}
		\begin{axis}[
			font=\scriptsize,
			xlabel={noise ($\%$)},
			ylabel={Corr $(\uparrow)$},
			ymin=64,
			ytick={64, 66, ..., 74},
			legend pos=south west,
			width=1.1\linewidth,
			height=1.1\linewidth,
		]

		\addplot[color=CB91_Violet, mark=*]
		coordinates {
			(5, 74.06)
			(10, 73.28)
			(15, 72.40)
			(20, 71.64)
			(25, 70.25)
			(30, 69.50)
			(35, 67.84)
			(40, 66.85)
		};
		\addlegendentry{Self-MM}
		\addplot[color=my_Green, mark=*]
		coordinates {
			(5, 74.70)
			(10, 73.92)
			(15, 73.05)
			(20, 72.29)
			(25, 70.91)
			(30, 70.16)
			(35, 68.52)
			(40, 67.51)
		};
		\addlegendentry{Self-MM+\ours}
		\end{axis}
	\end{tikzpicture}
\end{subfigure}
\caption{\emph{Modality dominance} analysis on MOSEI. Text modality input completely dropped with probability ranging from 5\% to 40\% or replaced with a mean representation over the training set. Average metrics reported over the two scenarios. $\uparrow/\downarrow$: higher/lower is better.}
\label{fig:dom_mult_selfmm}
\end{figure*}


\subsubsection{Modality dominance}

For dominance analysis, we inject noise solely into the text modality to assess the conditional dependence of the model on language to make predictions. The noise is applied in two forms: completely dropping the text modality input with probability $p$ or replacing it with a mean representation
over the training set~\cite{wu2022characterizing}.
We average the results across these two noise variants.

\autoref{fig:dom_mult_selfmm} shows the results for both MulT and Self-MM models. It reveals that text-only noise significantly affects both models, confirming the text dominance in MSA models as observed in prior work ~\cite{georgiou21_interspeech, hazarika2022analyzing}. Importantly, \ours-trained models consistently outperform the baselines across all examined noise levels.

\begin{figure}
\centering
\begin{subfigure}{0.9\columnwidth}
\centering
\begin{tikzpicture}
\begin{axis}[
	font=\scriptsize,
	xlabel={data percentage $(\%)$},
	ylabel={Acc-$2$ $(\uparrow)$},
	ymin=80,
	ytick={80, 81, ..., 85},
	legend pos=south east, 
	width=.9\columnwidth,
	height=.6\columnwidth,
]

\addplot[color=CB91_Violet, mark=*] coordinates {
	(10, 80.35)
	(20, 81.32)
	(30, 81.83)
	(50, 82.93)
	(80, 83.41)
	(100, 84.07)
};
\addlegendentry{MulT}

\addplot[color=my_Green, mark=*] coordinates {
	(10, 81.77)
	(20, 82.25)
	(30, 82.80)
	(50, 83.67)
	(80, 84.02)
	(100, 84.44)
};
\addlegendentry{MulT+\ours}



\end{axis}
\end{tikzpicture}
\end{subfigure}
\begin{subfigure}[h]{0.9\columnwidth}
\centering
\begin{tikzpicture}
\begin{axis}[
	font=\scriptsize,
	xlabel={data percentage $(\%)$},
	ylabel={MAE $(\downarrow)$},
	ymin=55,
	ytick={56, 58, ..., 64},
	legend pos=north east, 
	width=.9\columnwidth,
	height=.6\columnwidth,
]

\addplot[color=CB91_Violet, mark=*] coordinates {
	(10, 65.15)
	(20, 62.47)
	(30, 60.26)
	(50, 58.8)
	(80, 57.05)
	(100, 56.37)
};
\addlegendentry{MulT}


\addplot[color=my_Green, mark=*] coordinates {
	(10, 62.37)
	(20, 61.00)
	(30, 59.05)
	(50, 58.09)
	(80, 56.71)
	(100, 55.9)
};
\addlegendentry{MulT+\ours}


\end{axis}
\end{tikzpicture}
\label{fig:mae_rates_sims}
\end{subfigure}
\caption{\emph{Limited data} analysis. Model trained on progressively larger portions of MOSEI. $\uparrow/\downarrow$: higher/lower is better.}
\label{fig:powmix_limited}
\end{figure}
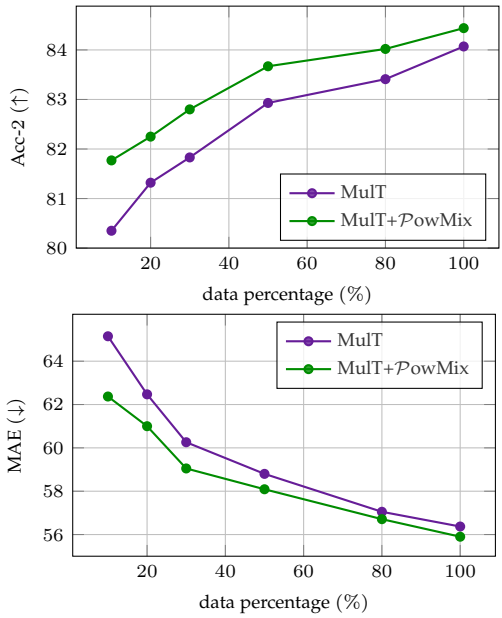

\subsubsection{Limited data}

This experiment investigates the effectiveness of \ours in limited data scenarios. We opt for the MulT model for its simple training process and lower parameter usage. The model is trained on progressively larger portions of MOSEI.
As shown in \autoref{fig:powmix_limited}, the model trained with \ours consistently outperforms the baseline across all data sizes. This performance enhancement is most pronounced in the lower data regime, specifically between $10-20\%$ of the data, where we observe an improvement of $\approx 1.21$ Acc-$2$ and $\approx 0.022$ MAE. This finding underlines the similarity of \ours to other regularization techniques, often showing greater improvements in scenarios with limited data.

%% file: tex/conclusion.tex
\section{Conclusions}
\label{sec:conclusion}

The increasing complexity of neural networks, especially in multimodal scenarios, underscores the critical need for effective regularization techniques. With the focus of MSA research on developing advanced architectures and diverse learning strategies, there is a clear demand for versatile multimodal regularization methods. To address this, we have introduced \ours, a novel approach specifically tailored for multimodal tasks. \ours incorporates five key elements: 1) generating a varying number of mixed examples, 2) mixing factor reweighting, 3) anisotropic mixing, 4) dynamic mixing, and 5) cross-modal label mixing. These elements collectively form an algorithm that improves training in multimodal contexts.

Our extensive experimentats across various MSA datasets and models demonstrate the broad applicability and consistent performance improvements of \ours. Detailed ablation studies uncover the synergistic nature of its components, emphasizing that the full set of components is essential for its effective operation. Removing any algorithmic component results in performance degradation. Moreover, the ablation highlights the robustness of the algorithm to several hyperparameter choices, a valuable quality of the algorithm. Extented analysis shows that \ours operates differently yet effectively in both early and late fusion architectures. Moreover, we find that the integration of \ours into MSA models preserves robustness without sacrificing performance or enhancing text dominance and offers consistent gains across different scales of data.

Future research on \ours is promising. One important direction is the application of \ours in other multimodal tasks beyond MSA, to further validate its versatility and efficacy in diverse environments. Investigating the integration of \ours with more and diverse neural network architectures could further establish it as a generic multimodal regularizer. Another potential direction is to examine how \ours performs under unsupervised or semi-supervised learning scenarios. This could contribute to the development of more generalized multimodal learning frameworks, capable of handling a wider spectrum of real-world tasks. Further in-depth analysis of each component of \ours could provide a clearer understanding of both the individual as well as the collective contributions to the learning process. Such insights could lead to the development of more refined, targeted and even explicit regularization methods. Finally, the exploration of learnable mixing strategies could further advance the state of the current multimodal regularization arsenal.

%% file: powmix.bbl
\begin{thebibliography}{10}
\providecommand{\url}[1]{#1}
\csname url@samestyle\endcsname
\providecommand{\newblock}{\relax}
\providecommand{\bibinfo}[2]{#2}
\providecommand{\BIBentrySTDinterwordspacing}{\spaceskip=0pt\relax}
\providecommand{\BIBentryALTinterwordstretchfactor}{4}
\providecommand{\BIBentryALTinterwordspacing}{\spaceskip=\fontdimen2\font plus
\BIBentryALTinterwordstretchfactor\fontdimen3\font minus \fontdimen4\font\relax}
\providecommand{\BIBforeignlanguage}[2]{{%
\expandafter\ifx\csname l@#1\endcsname\relax
\typeout{** WARNING: IEEEtran.bst: No hyphenation pattern has been}%
\typeout{** loaded for the language `#1'. Using the pattern for}%
\typeout{** the default language instead.}%
\else
\language=\csname l@#1\endcsname
\fi
#2}}
\providecommand{\BIBdecl}{\relax}
\BIBdecl

\bibitem{narayanan_2013_behavioral_signal}
S.~Narayanan and P.~G. Georgiou, ``Behavioral signal processing: Deriving human behavioral informatics from speech and language,'' \emph{Proceedings of the IEEE}, vol. 101, no.~5, pp. 1203--1233, 2013.

\bibitem{filby_2019_child_robot}
P.~P. Filntisis, N.~Efthymiou, P.~Koutras, G.~Potamianos, and P.~Maragos, ``Fusing body posture with facial expressions for joint recognition of affect in child–robot interaction,'' \emph{IEEE Robotics and Automation Letters}, vol.~4, no.~4, pp. 4011--4018, 2019.

\bibitem{poria2017review}
S.~Poria, E.~Cambria, R.~Bajpai, and A.~Hussain, ``A review of affective computing: From unimodal analysis to multimodal fusion,'' \emph{Information Fusion}, vol.~37, pp. 98--125, 2017.

\bibitem{stappen2021multimodal}
L.~Stappen, A.~Baird, L.~Schumann, and S.~Bjorn, ``The multimodal sentiment analysis in car reviews (muse-car) dataset: Collection, insights and improvements,'' \emph{IEEE Transactions on Affective Computing}, 2021.

\bibitem{affect_in_education}
E.~Yadegaridehkordi, N.~F. B.~M. Noor, M.~N.~B. Ayub, H.~B. Affal, and N.~B. Hussin, ``Affective computing in education: A systematic review and future research,'' \emph{Computers \& Education}, vol. 142, p. 103649, 2019.

\bibitem{metallinou2012context}
A.~Metallinou, M.~Wollmer, A.~Katsamanis, F.~Eyben, B.~Schuller, and S.~Narayanan, ``Context-sensitive learning for enhanced audiovisual emotion classification,'' \emph{IEEE Transactions on Affective Computing}, vol.~3, no.~2, pp. 184--198, 2012.

\bibitem{soleymani2017survey}
M.~Soleymani, D.~Garcia, B.~Jou, B.~Schuller, S.-F. Chang, and M.~Pantic, ``A survey of multimodal sentiment analysis,'' \emph{Image and Vision Computing}, vol.~65, pp. 3--14, 2017.

\bibitem{zadeh2017tensor}
A.~Zadeh, M.~Chen, S.~Poria, E.~Cambria, and L.-P. Morency, ``Tensor fusion network for multimodal sentiment analysis,'' in \emph{Proceedings of the 2017 Conference on Empirical Methods in Natural Language Processing}, 2017, pp. 1103--1114.

\bibitem{tsai2019multimodal}
Y.-H.~H. Tsai, S.~Bai, P.~P. Liang, J.~Z. Kolter, L.-P. Morency, and R.~Salakhutdinov, ``Multimodal transformer for unaligned multimodal language sequences,'' in \emph{Proceedings of the 57th Annual Meeting of the Association for Computational Linguistics}, 2019, pp. 6558--6569.

\bibitem{hazarika2020misa}
D.~Hazarika, R.~Zimmermann, and S.~Poria, ``Misa: Modality-invariant and-specific representations for multimodal sentiment analysis,'' in \emph{Proceedings of the 28th ACM International Conference on Multimedia}, 2020, pp. 1122--1131.

\bibitem{yu2021learning}
W.~Yu, H.~Xu, Z.~Yuan, and J.~Wu, ``Learning modality-specific representations with self-supervised multi-task learning for multimodal sentiment analysis,'' in \emph{Proceedings of the AAAI Conference on Artificial Intelligence}, vol.~35, no.~12, 2021, pp. 10\,790--10\,797.

\bibitem{sun2022learning}
Y.~Sun, S.~Mai, and H.~Hu, ``Learning to learn better unimodal representations via adaptive multimodal meta-learning,'' \emph{IEEE Transactions on Affective Computing}, 2022.

\bibitem{hessel-lee-2020-multimodal}
J.~Hessel and L.~Lee, ``Does my multimodal model learn cross-modal interactions? it{'}s harder to tell than you might think!'' in \emph{Proceedings of the 2020 Conference on Empirical Methods in Natural Language Processing (EMNLP)}.\hskip 1em plus 0.5em minus 0.4em\relax Online: Association for Computational Linguistics, Nov. 2020, pp. 861--877.

\bibitem{huang_2021_mm_better}
Y.~Huang, C.~Du, Z.~Xue, X.~Chen, H.~Zhao, and L.~Huang, ``What makes multi-modal learning better than single (provably),'' \emph{Advances in Neural Information Processing Systems}, vol.~34, pp. 10\,944--10\,956, 2021.

\bibitem{wang2020makes}
W.~Wang, D.~Tran, and M.~Feiszli, ``What makes training multi-modal classification networks hard?'' in \emph{Proceedings of the IEEE/CVF conference on computer vision and pattern recognition}, 2020, pp. 12\,695--12\,705.

\bibitem{wu2022characterizing}
N.~Wu, S.~Jastrzebski, K.~Cho, and K.~J. Geras, ``Characterizing and overcoming the greedy nature of learning in multi-modal deep neural networks,'' in \emph{International Conference on Machine Learning, {ICML} 2022, 17-23 July 2022, Baltimore, Maryland, {USA}}, ser. Proceedings of Machine Learning Research, vol. 162.\hskip 1em plus 0.5em minus 0.4em\relax {PMLR}, 2022, pp. 24\,043--24\,055.

\bibitem{GKOUMAS2021WhatMakesTheDiff}
D.~Gkoumas, Q.~Li, C.~Lioma, Y.~Yu, and D.~Song, ``What makes the difference? an empirical comparison of fusion strategies for multimodal language analysis,'' \emph{Information Fusion}, vol.~66, pp. 184--197, 2021.

\bibitem{huang_22_mm_fail}
Y.~Huang, J.~Lin, C.~Zhou, H.~Yang, and L.~Huang, ``Modality competition: What makes joint training of multi-modal network fail in deep learning? ({P}rovably),'' in \emph{Proceedings of the 39th International Conference on Machine Learning}, ser. Proceedings of Machine Learning Research, vol. 162.\hskip 1em plus 0.5em minus 0.4em\relax PMLR, 17--23 Jul 2022, pp. 9226--9259.

\bibitem{du2023onunimodal}
C.~Du, J.~Teng, T.~Li, Y.~Liu, T.~Yuan, Y.~Wang, Y.~Yuan, and H.~Zhao, ``On uni-modal feature learning in supervised multi-modal learning,'' in \emph{International Conference on Machine Learning, {ICML} 2023, 23-29 July 2023, Honolulu, Hawaii, {USA}}, ser. Proceedings of Machine Learning Research, vol. 202.\hskip 1em plus 0.5em minus 0.4em\relax {PMLR}, 2023, pp. 8632--8656.

\bibitem{liu2023learning}
Z.~Liu, Z.~Tang, X.~Shi, A.~Zhang, M.~Li, A.~Shrivastava, and A.~G. Wilson, ``Learning multimodal data augmentation in feature space,'' in \emph{The Eleventh International Conference on Learning Representations, {ICLR} 2023, Kigali, Rwanda, May 1-5, 2023}.\hskip 1em plus 0.5em minus 0.4em\relax OpenReview.net, 2023.

\bibitem{georgiou21_interspeech}
E.~Georgiou, G.~Paraskevopoulos, and A.~Potamianos, ``{M3: MultiModal Masking Applied to Sentiment Analysis},'' in \emph{Proc. Interspeech 2021}, 2021, pp. 2876--2880.

\bibitem{liu2022_sims2}
Y.~Liu, Z.~Yuan, H.~Mao, Z.~Liang, W.~Yang, Y.~Qiu, T.~Cheng, X.~Li, H.~Xu, and K.~Gao, ``Make acoustic and visual cues matter: Ch-sims v2. 0 dataset and av-mixup consistent module,'' in \emph{Proceedings of the 2022 International Conference on Multimodal Interaction}, 2022, pp. 247--258.

\bibitem{zhang2018mixup}
H.~Zhang, M.~Ciss{\'{e}}, Y.~N. Dauphin, and D.~Lopez{-}Paz, ``mixup: Beyond empirical risk minimization,'' in \emph{6th International Conference on Learning Representations, {ICLR} 2018}.\hskip 1em plus 0.5em minus 0.4em\relax OpenReview.net, 2018.

\bibitem{transmix}
J.-N. Chen, S.~Sun, J.~He, P.~Torr, A.~Yuille, and S.~Bai, ``Transmix: Attend to mix for vision transformers,'' in \emph{The IEEE Conference on Computer Vision and Pattern Recognition (CVPR)}, June 2022.

\bibitem{venkataramanan2023Embedding}
S.~Venkataramanan, E.~Kijak, L.~Amsaleg, and Y.~Avrithis, ``Embedding space interpolation beyond mini-batch, beyond pairs and beyond examples,'' in \emph{Advances in neural information processing systems}, 2023.

\bibitem{zadeh2016multimodal}
A.~Zadeh, R.~Zellers, E.~Pincus, and L.-P. Morency, ``Multimodal sentiment intensity analysis in videos: Facial gestures and verbal messages,'' \emph{IEEE Intelligent Systems}, vol.~31, no.~6, pp. 82--88, 2016.

\bibitem{zadeh2018multimodal}
A.~Zadeh and P.~Pu, ``Multimodal language analysis in the wild: Cmu-mosei dataset and interpretable dynamic fusion graph,'' in \emph{Proceedings of the 56th Annual Meeting of the Association for Computational Linguistics (Long Papers)}, 2018.

\bibitem{yu2020ch}
W.~Yu, H.~Xu, F.~Meng, Y.~Zhu, Y.~Ma, J.~Wu, J.~Zou, and K.~Yang, ``Ch-sims: A chinese multimodal sentiment analysis dataset with fine-grained annotation of modality,'' in \emph{Proceedings of the 58th Annual Meeting of the Association for Computational Linguistics}, 2020, pp. 3718--3727.

\bibitem{poria2017multi}
S.~Poria, E.~Cambria, D.~Hazarika, N.~Mazumder, A.~Zadeh, and L.-P. Morency, ``Multi-level multiple attentions for contextual multimodal sentiment analysis,'' in \emph{2017 IEEE International Conference on Data Mining (ICDM)}.\hskip 1em plus 0.5em minus 0.4em\relax IEEE, 2017, pp. 1033--1038.

\bibitem{gu2018multimodal}
Y.~Gu, K.~Yang, S.~Fu, S.~Chen, X.~Li, and I.~Marsic, ``Multimodal affective analysis using hierarchical attention strategy with word-level alignment,'' in \emph{Proceedings of the conference. Association for Computational Linguistics. Meeting}, vol. 2018.\hskip 1em plus 0.5em minus 0.4em\relax NIH Public Access, 2018, p. 2225.

\bibitem{georgiou2019deep}
E.~Georgiou, C.~Papaioannou, and A.~Potamianos, ``Deep hierarchical fusion with application in sentiment analysis,'' \emph{Interspeech 2019}, 2019.

\bibitem{tsai20routing}
Y.~H. Tsai, M.~Ma, M.~Yang, R.~Salakhutdinov, and L.~Morency, ``Multimodal routing: Improving local and global interpretability of multimodal language analysis,'' in \emph{Proc. of the 2020 Conference on Empirical Methods in Natural Language Processing, {EMNLP}}, 2020, pp. 1823--1833.

\bibitem{joshi-etal-2022-cogmen}
A.~Joshi, A.~Bhat, A.~Jain, A.~Singh, and A.~Modi, ``{COGMEN}: {CO}ntextualized {GNN} based multimodal emotion recognitio{N},'' in \emph{Proceedings of the 2022 Conference of the North American Chapter of the Association for Computational Linguistics: Human Language Technologies}.\hskip 1em plus 0.5em minus 0.4em\relax Association for Computational Linguistics, 2022, pp. 4148--4164.

\bibitem{rahman2020integrating}
W.~Rahman, M.~K. Hasan, S.~Lee, A.~B. Zadeh, C.~Mao, L.-P. Morency, and E.~Hoque, ``Integrating multimodal information in large pretrained transformers,'' in \emph{Proceedings of the 58th Annual Meeting of the Association for Computational Linguistics}, 2020, pp. 2359--2369.

\bibitem{devlin2019bert}
J.~Devlin, M.-W. Chang, K.~Lee, and K.~Toutanova, ``{BERT}: Pre-training of deep bidirectional transformers for language understanding,'' in \emph{Proceedings of the 2019 Conference of the North {A}merican Chapter of the Association for Computational Linguistics: Human Language Technologies, Volume 1 (Long and Short Papers)}.\hskip 1em plus 0.5em minus 0.4em\relax Association for Computational Linguistics, Jun. 2019, pp. 4171--4186.

\bibitem{pham2019found}
H.~Pham, P.~P. Liang, T.~Manzini, L.-P. Morency, and B.~P{\'o}czos, ``Found in translation: Learning robust joint representations by cyclic translations between modalities,'' in \emph{Proceedings of the AAAI Conference on Artificial Intelligence}, vol.~33, no.~01, 2019, pp. 6892--6899.

\bibitem{han2021improving}
W.~Han, H.~Chen, and S.~Poria, ``Improving multimodal fusion with hierarchical mutual information maximization for multimodal sentiment analysis,'' in \emph{Proceedings of the 2021 Conference on Empirical Methods in Natural Language Processing}, 2021, pp. 9180--9192.

\bibitem{sun2023efficient}
L.~Sun, Z.~Lian, B.~Liu, and J.~Tao, ``Efficient multimodal transformer with dual-level feature restoration for robust multimodal sentiment analysis,'' \emph{IEEE Transactions on Affective Computing}, 2023.

\bibitem{hu-etal-2022-unimse}
G.~Hu, T.-E. Lin, Y.~Zhao, G.~Lu, Y.~Wu, and Y.~Li, ``{U}ni{MSE}: Towards unified multimodal sentiment analysis and emotion recognition,'' in \emph{Proceedings of the 2022 Conference on Empirical Methods in Natural Language Processing}.\hskip 1em plus 0.5em minus 0.4em\relax Association for Computational Linguistics, Dec. 2022, pp. 7837--7851.

\bibitem{RAFFEL2020T5}
C.~Raffel, N.~Shazeer, A.~Roberts, K.~Lee, S.~Narang, M.~Matena, Y.~Zhou, W.~Li, and P.~J. Liu, ``Exploring the limits of transfer learning with a unified text-to-text transformer,'' \emph{Journal of Machine Learning Research}, vol.~21, no. 140, pp. 1--67, 2020.

\bibitem{Goodfellow-et-al-2016}
I.~Goodfellow, Y.~Bengio, and A.~Courville, \emph{Deep Learning}.\hskip 1em plus 0.5em minus 0.4em\relax MIT Press, 2016.

\bibitem{yun2019cutmix}
S.~Yun, D.~Han, S.~Chun, S.~J. Oh, Y.~Yoo, and J.~Choe, ``Cutmix: Regularization strategy to train strong classifiers with localizable features,'' in \emph{2019 {IEEE/CVF} International Conference on Computer Vision, {ICCV} 2019, Seoul, Korea (South), October 27 - November 2, 2019}.\hskip 1em plus 0.5em minus 0.4em\relax {IEEE}, 2019, pp. 6022--6031.

\bibitem{liu2022automix}
Z.~Liu, S.~Li, D.~Wu, Z.~Chen, L.~Wu, J.~Guo, and S.~Z. Li, ``Automix: Unveiling the power of mixup for stronger classifiers,'' in \emph{European Conference on Computer Vision}, 2022, pp. 441--458.

\bibitem{li2022openmixup}
S.~Li, Z.~Wang, Z.~Liu, D.~Wu, and S.~Z. Li, ``Openmixup: Open mixup toolbox and benchmark for visual representation learning,'' \emph{arXiv preprint arXiv:2209.04851}, 2022.

\bibitem{yoon2021ssmix}
S.~Yoon, G.~Kim, and K.~Park, ``Ssmix: Saliency-based span mixup for text classification,'' in \emph{Findings of the Association for Computational Linguistics: ACL-IJCNLP 2021}, 2021, pp. 3225--3234.

\bibitem{kim21c_interspeech}
G.~Kim, D.~K. Han, and H.~Ko, ``{SpecMix : A Mixed Sample Data Augmentation Method for Training with Time-Frequency Domain Features},'' in \emph{Proc. Interspeech 2021}, 2021, pp. 546--550.

\bibitem{verma2019manifold}
V.~Verma, A.~Lamb, C.~Beckham, A.~Najafi, I.~Mitliagkas, D.~Lopez-Paz, and Y.~Bengio, ``Manifold mixup: Better representations by interpolating hidden states,'' in \emph{International conference on machine learning}.\hskip 1em plus 0.5em minus 0.4em\relax PMLR, 2019, pp. 6438--6447.

\bibitem{Guo_2020_nonlinear}
H.~Guo, ``Nonlinear mixup: Out-of-manifold data augmentation for text classification,'' \emph{Proceedings of the AAAI Conference on Artificial Intelligence}, vol.~34, no.~04, pp. 4044--4051, Apr. 2020.

\bibitem{chou_2020_remix}
H.-P. Chou, S.-C. Chang, J.-Y. Pan, W.~Wei, and D.-C. Juan, ``Remix: Rebalanced mixup,'' in \emph{Computer Vision – ECCV 2020 Workshops, 2020, Proceedings, Part VI}, 2020, p. 95–110.

\bibitem{jindal20_speechmix}
A.~Jindal, N.~E. Ranganatha, A.~Didolkar, A.~G. Chowdhury, D.~Jin, R.~Sawhney, and R.~R. Shah, ``{SpeechMix — Augmenting Deep Sound Recognition Using Hidden Space Interpolations},'' in \emph{Proc. Interspeech 2020}, 2020, pp. 861--865.

\bibitem{sun_2020_mixuptrans}
L.~Sun, C.~Xia, W.~Yin, T.~Liang, P.~Yu, and L.~He, ``Mixup-transformer: Dynamic data augmentation for {NLP} tasks,'' in \emph{Proceedings of the 28th International Conference on Computational Linguistics}.\hskip 1em plus 0.5em minus 0.4em\relax International Committee on Computational Linguistics, 2020, pp. 3436--3440.

\bibitem{Hao_2023_WACV}
X.~Hao, Y.~Zhu, S.~Appalaraju, A.~Zhang, W.~Zhang, B.~Li, and M.~Li, ``Mixgen: A new multi-modal data augmentation,'' in \emph{Proceedings of the IEEE/CVF Winter Conference on Applications of Computer Vision (WACV) Workshops}, January 2023, pp. 379--389.

\bibitem{wang_22_cmc}
T.~Wang, W.~Jiang, Z.~Lu, F.~Zheng, R.~Cheng, C.~Yin, and P.~Luo, ``{VLM}ixer: Unpaired vision-language pre-training via cross-modal {C}ut{M}ix,'' in \emph{Proceedings of the 39th International Conference on Machine Learning}, ser. Proceedings of Machine Learning Research, vol. 162.\hskip 1em plus 0.5em minus 0.4em\relax PMLR, 17--23 Jul 2022, pp. 22\,680--22\,690.

\bibitem{mao-etal-2022-sena}
H.~Mao, Z.~Yuan, H.~Xu, W.~Yu, Y.~Liu, and K.~Gao, ``{M}-{SENA}: An integrated platform for multimodal sentiment analysis,'' in \emph{Proceedings of the 60th Annual Meeting of the Association for Computational Linguistics: System Demonstrations}.\hskip 1em plus 0.5em minus 0.4em\relax Association for Computational Linguistics, 2022, pp. 204--213.

\bibitem{wolf2020transformers}
T.~Wolf, L.~Debut, V.~Sanh, J.~Chaumond, C.~Delangue, A.~Moi, P.~Cistac, T.~Rault, R.~Louf, M.~Funtowicz \emph{et~al.}, ``Transformers: State-of-the-art natural language processing,'' in \emph{Proceedings of the 2020 conference on empirical methods in natural language processing: system demonstrations}, 2020, pp. 38--45.

\bibitem{degottex2014covarep}
G.~Degottex, J.~Kane, T.~Drugman, T.~Raitio, and S.~Scherer, ``Covarep—a collaborative voice analysis repository for speech technologies,'' in \emph{2014 ieee international conference on acoustics, speech and signal processing (icassp)}.\hskip 1em plus 0.5em minus 0.4em\relax IEEE, 2014, pp. 960--964.

\bibitem{mcfee2015librosa}
B.~McFee, C.~Raffel, D.~Liang, D.~P. Ellis, M.~McVicar, E.~Battenberg, and O.~Nieto, ``librosa: Audio and music signal analysis in python,'' in \emph{Proceedings of the 14th python in science conference}, vol.~8.\hskip 1em plus 0.5em minus 0.4em\relax Citeseer, 2015, pp. 18--25.

\bibitem{baltrusaitis2018openface}
T.~Baltrusaitis, A.~Zadeh, Y.~C. Lim, and L.-P. Morency, ``Openface 2.0: Facial behavior analysis toolkit,'' in \emph{2018 13th IEEE international conference on automatic face \& gesture recognition (FG 2018)}.\hskip 1em plus 0.5em minus 0.4em\relax IEEE, 2018, pp. 59--66.

\bibitem{jin2023weakening}
C.~Jin, C.~Luo, M.~Yan, G.~Zhao, G.~Zhang, and S.~Zhang, ``Weakening the dominant role of text: Cmosi dataset and multimodal semantic enhancement network,'' \emph{IEEE Transactions on Neural Networks and Learning Systems}, pp. 1--15, 2023.

\bibitem{cambria2018benchmarking}
E.~Cambria, D.~Hazarika, S.~Poria, A.~Hussain, and R.~Subramanyam, ``Benchmarking multimodal sentiment analysis,'' in \emph{Computational Linguistics and Intelligent Text Processing: 18th International Conference, CICLing 2017, Budapest, Hungary, April 17--23, 2017, Revised Selected Papers, Part II 18}.\hskip 1em plus 0.5em minus 0.4em\relax Springer, 2018, pp. 166--179.

\bibitem{paszke2019pytorch}
A.~Paszke, S.~Gross, F.~Massa, A.~Lerer, J.~Bradbury, G.~Chanan, T.~Killeen, Z.~Lin, N.~Gimelshein, L.~Antiga \emph{et~al.}, ``Pytorch: An imperative style, high-performance deep learning library,'' \emph{Advances in neural information processing systems}, vol.~32, 2019.

\bibitem{kingma2014adam}
D.~P. Kingma and J.~Ba, ``Adam: A method for stochastic optimization,'' \emph{arXiv preprint arXiv:1412.6980}, 2014.

\bibitem{jing2020self}
L.~Jing and Y.~Tian, ``Self-supervised visual feature learning with deep neural networks: A survey,'' \emph{IEEE transactions on pattern analysis and machine intelligence}, vol.~43, no.~11, pp. 4037--4058, 2020.

\bibitem{hazarika2022analyzing}
D.~Hazarika, Y.~Li, B.~Cheng, S.~Zhao, R.~Zimmermann, and S.~Poria, ``Analyzing modality robustness in multimodal sentiment analysis,'' \emph{arXiv preprint arXiv:2205.15465}, 2022.

\end{thebibliography}
